%% file: main.tex
\pdfoutput=1

\documentclass[10pt,twocolumn,letter]{article}

\usepackage{amssymb}
\usepackage[pagenumbers]{cvpr} 
\usepackage{lineno}
\usepackage{wasysym}
\usepackage{marvosym}


\definecolor{cvprblue}{rgb}{0.21,0.49,0.74}
\usepackage[pagebackref,breaklinks,colorlinks,allcolors=cvprblue]{hyperref}

\usepackage{multirow}
\usepackage{tcolorbox}
\usepackage{wrapfig}

\def\paperID{14995} 
\def\confName{CVPR}
\def\confYear{2025}

\title{Automatic Evaluation for Text-to-image Generation: Task-decomposed Framework, Distilled Training, and Meta-evaluation Benchmark}

\author{Rong-Cheng Tu$^{1*}$ \quad Zi-Ao Ma$^{1*}$ \quad Tian Lan$^{1*}$ \quad Yuehao Zhao$^{1}$\thanks{\quad Equal contributions} \\
Heyan Huang$^{1}$ \quad Xian-Ling Mao$^{1}$\thanks{\quad Corresponding author}\\
$^{1}$School of Computer Science and Technology, Beijing Institute of Technology\\
{\texttt{\{turongcheng,lantiangmftby\}@gmail.com}} \quad {\texttt{maoxl@bit.edu.cn}}\\
\url{https://github.com/maziao/T2I-Eval}
}

\begin{document}
\maketitle
\input{sec/0_abstract}    
\input{sec/1_introduction}

\input{sec/2_relatedwork}

\input{sec/3_method}
\input{sec/4_expriments}
{
    \small
    \bibliographystyle{ieeenat_fullname}
    \bibliography{main}
}

\input{sec/appendix}

\end{document}

%% file: sec/0_abstract.tex
\begin{abstract}
Driven by the remarkable progress in diffusion models, text-to-image generation has made significant strides, creating a pressing demand for automatic quality evaluation of generated images. Current state-of-the-art automatic evaluation methods heavily rely on Multi-modal Large Language Models (MLLMs), particularly powerful commercial models like GPT-4o. While these models are highly effective, their substantial costs limit scalability in large-scale evaluations. Adopting open-source MLLMs is an alternative; however, their performance falls short due to significant limitations in processing multi-modal data compared to commercial MLLMs. 
To tackle these problems, we first propose a task decomposition evaluation framework based on GPT-4o to automatically construct a new training dataset, where the complex evaluation task is decoupled into simpler sub-tasks, effectively reducing the learning complexity. Based on this dataset, we design innovative training strategies to effectively distill GPT-4o's evaluation capabilities into a 7B open-source MLLM, MiniCPM-V-2.6. 
Furthermore, to reliably and comprehensively assess prior works and our proposed model, we manually annotate a meta-evaluation benchmark that includes chain-of-thought explanations alongside quality scores for generated images.
Experimental results demonstrate that our distilled open-source MLLM significantly outperforms the current state-of-the-art GPT-4o-base baseline, VIEScore, with over 4.6\% improvement in Spearman and Kendall correlations with human judgments. 
\end{abstract}

%% file: sec/1_introduction.tex
\section{Introduction}\label{sec:introduction}

The rapid advancements in diffusion models have significantly driven the progress of text-to-image generation models \cite{song2022denoisingdiffusionimplicitmodels,ho2020denoisingdiffusionprobabilisticmodels,rombach2022highresolutionimagesynthesislatent,podell2023sdxlimprovinglatentdiffusion,esser2024scalingrectifiedflowtransformers,peebles2023scalablediffusionmodelstransformers,ramesh2021zeroshottexttoimagegeneration,ramesh2022hierarchicaltextconditionalimagegeneration,li2024playgroundv25insightsenhancing,liu2024playgroundv3improvingtexttoimage}. While these models demonstrate the capability to generate highly creative visual content, the generated images often suffer from issues such as distorted appearances of major entities and incorrect alignment with the input text prompt \cite{cao2024survey,cao2024controllable,wan2024survey}. Automatically evaluating these issues can not only provide effective loss functions for training generative models to enhance their performance but also filter out low-quality generated images during inference, thereby improving user experience~\cite{stiennon2022learningsummarizehumanfeedback,nakano2022webgptbrowserassistedquestionansweringhuman}. Consequently, there is an urgent need for precise and automatic evaluation methods to assess the quality and fidelity of generated images~\cite{ku2023viescore,lu2023llmscore}.

To meet this need, early works like CLIP-based and BLIP-based scoring methods~\cite{radford2021learning} have been used to evaluate the semantic alignment between input text and generated images, yet they still have limitations in handling complex semantic relationships \cite{ku2023viescore}.
Recently, pre-trained Multi-modal Large Language Models (MLLMs) \cite{dong2024internlmxcomposer2masteringfreeformtextimage,hu2024minicpmunveilingpotentialsmall,wang2024qwen2vlenhancingvisionlanguagemodels,wu2024nextgptanytoanymultimodalllm} have demonstrated powerful semantic understanding and reasoning capabilities, exhibiting higher correlation with human judgments~\cite{ku2023viescore,lu2023llmscore,wiles2024revisitingtexttoimageevaluationgecko}. This has promoted researchers to develop MLLM-based automatic evaluation methods. These methods typically employ simple prompts, asking MLLMs to directly assess the quality of generated images by implicitly completing multiple complex judgment tasks. However, due to the overly simplistic prompt design, the models need exceptionally advanced semantic understanding and reasoning abilities to perform the evaluation on text-to-image generation task accurately.
Consequently, these methods often rely on advanced commercial models like GPT-4o \cite{achiam2023gpt} to achieve superior evaluation performance. The high costs associated with these models limit their applicability in large-scale evaluations. Adopting open-source MLLMs is an alternative, but due to their relatively weaker semantic understanding and reasoning abilities, they struggle to effectively handle complex evaluation tasks, resulting in poor evaluation performance. 

In light of this, we aim to enhance the capability of open-source MLLMs in  evaluating the quality of generated images. We argue that by decomposing the complex evaluation task into a series of simpler or fine-grained sub-tasks, open-source models can progressively complete them and accurately evaluate the qualities of generated images.

To this end, we propose a novel task-decomposed evaluation framework based on GPT-4o to automatically construct a training dataset to optimize open-source MLLMs for better evaluation performance.
Specifically, this framework first adopts GPT-4o to extract entities and their intrinsic properties, and relational attributes from the input text prompt. These extracted details are used to formulate questions for detailed evaluation across three dimensions: visual appearance, intrinsic properties, and relational attributes. Next, GPT-4o answers each question based on the image and its caption, comparing the response with the ground-truth extracted from the input text to produce detailed explanations and quality scores. For each evaluation dimension, we aggregate all predicted results for the questions to provide corresponding explanations and score the dimension's quality. Finally, by considering all evaluated dimensions, the framework delivers an overall judgment.

Based on the training dataset automatically constructed through the aforementioned framework, we propose a novel and practical paradigm to fine-tune the 7B open-source MLLM, MiniCPM-V-2.6, into a highly efficient automatic evaluation model. 
Additionally, to comprehensively and reliably evaluate the performance of existing baselines and our fine-tuned model, we manually annotate a meta-evaluation benchmark, which also evaluates the generated images from visual appearance, intrinsic properties and relational attributes~\cite{lan2024criticevalevaluatinglargelanguage,lan2024traininglanguagemodelscritique}. The fine-tuned model, training dataset and meta-evaluation benchmark are openly available. 
In a nutshell, our contributions are four-fold:
\begin{itemize}
    \item  We propose a fine-grained automatic evaluation framework that decomposes complex evaluation tasks into simpler sub-tasks. It is used to construct a high-quality training dataset, reducing the learning difficulty for open-source MLLMs.
    \item Based on the training dataset, we propose novel training strategies to effectively optimize the the 7B open-source MLLM, MiniCPM-V-2.6, transforming it into a remarkable image quality evaluation model.
    \item  We manually annotate a test dataset, serving as a meta-evaluation benchmark for assessing the performance of existing evaluation methods and our distilled MLLM. 
    \item  Extensive experimental results validate the effectiveness and superiority of our proposed automatic evaluation framework and the fine-tuned evaluation model.
\end{itemize}

%% file: sec/2_relatedwork.tex
\section{Related Work}
\subsection{Image Generation}
In recent years, with the rapid advancement of diffusion models and large-scale image datasets \cite{young-etal-2014-image, lin2015microsoftcococommonobjects,karras2018progressivegrowinggansimproved,karras2019stylebasedgeneratorarchitecturegenerative}, text-to-image generation models \cite{rombach2022highresolutionimagesynthesislatent,podell2023sdxlimprovinglatentdiffusion,sun2024diffusion,shuai2024survey,esser2024scalingrectifiedflowtransformers,peebles2023scalablediffusionmodelstransformers,ramesh2021zeroshottexttoimagegeneration,ramesh2022hierarchicaltextconditionalimagegeneration}  have achieved remarkable progress. Pioneering works like DDPM \cite{ho2020denoisingdiffusionprobabilisticmodels} successfully trained diffusion models for image generation; Stable Diffusion \cite{rombach2022highresolutionimagesynthesislatent,podell2023sdxlimprovinglatentdiffusion} utilized latent diffusion models to generate high-resolution images; DiT \cite{peebles2023scalablediffusionmodelstransformers} adopted transformer as the backbone to construct diffusion models for high-quality images. Subsequently, an increasing number of transformer-based methods \cite{ramesh2021zeroshottexttoimagegeneration,ramesh2022hierarchicaltextconditionalimagegeneration,li2024playgroundv25insightsenhancing,esser2024scalingrectifiedflowtransformers,liu2024playgroundv3improvingtexttoimage} have been proposed to generate high-fidelity images.

While these models \cite{rombach2022highresolutionimagesynthesislatent,podell2023sdxlimprovinglatentdiffusion,esser2024scalingrectifiedflowtransformers,peebles2023scalablediffusionmodelstransformers,ramesh2021zeroshottexttoimagegeneration,ramesh2022hierarchicaltextconditionalimagegeneration} demonstrate the capability to generate highly creative images, the outputs still suffer from distorted major entities and misalignment with text prompts.  These limitations have spurred researchers to develop more precise and automated evaluation methods to assess both the quality of generated images and their correspondence to text prompts. 

\begin{figure*}[h]
    \centering
    \includegraphics[width=1.0\linewidth]{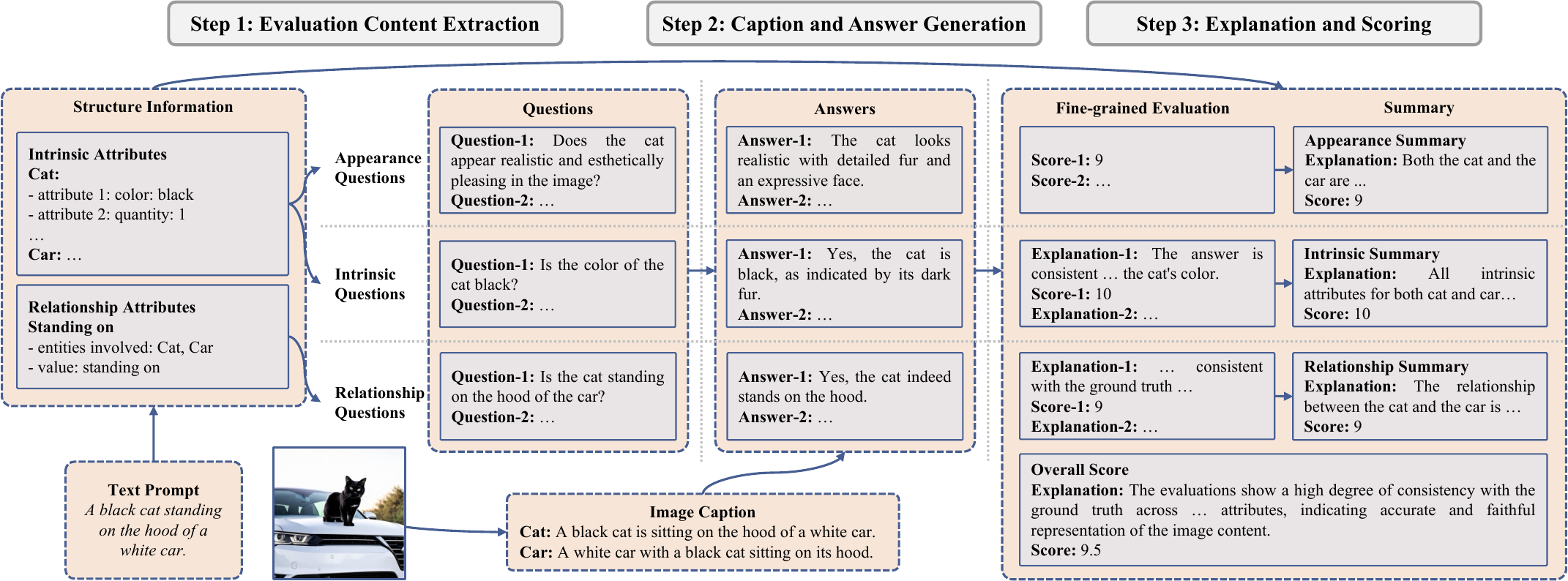}
    \caption{The overview of our proposed Task Decomposition Evaluation Framework, consisting of three steps: (1) Evaluation Content Extraction and Question Generation; (2) Caption and Answer Generation; (3) Explanation and Scoring.}
    \label{fig:framework_pipeline}
\end{figure*}

\subsection{Evaluation of Model-generated Images}
To automatically evaluate the quality of generated images, in the early years, the metrics Inception Score (IS)~\cite{salimans2016improved} and Fréchet Inception Distance (FID)~\cite{heusel2017gans} were proposed to assess the the clarity and diversity of generated images by comparing them to real images. 
Moreover, benefiting from the the powerful feature extracting capabilities of the CLIP~\cite{radford2021learning} and BLIP~\cite{li2022blip} models, the CLIP-based and BLIP-based scoring methods \cite{hessel2021clipscore,wu2023humanpreferencescorev2} measure the consistency between generated images and corresponding text prompts, but these metrics fail to assess the complex object-level alignment. To address this issue, visual-question-answering (VQA)-based methods~\cite{lin2024evaluatingtexttovisualgenerationimagetotext,wiles2024revisitingtexttoimageevaluationgecko,yarom2023readimprovingtextimagealignment} are proposed. VQA-based methods first decompose the text prompt into simple questions using Large Language Models (LLMs), and then evaluate the quality of generated images by computing the accuracy of the `yes/no' answers of these questions. 

Recently, there is an emerging trend to leverage the reasoning capabilities of  MLLMs, like GPT-4o, to directly assess the alignment between generated images and input text, exhibiting better correlation with human judgments and great interpretability~\cite{lu2023llmscore,ku2023viescore,tan2024evalalignsupervisedfinetuningmultimodal}. 
For example, VIEscore~\cite{ku2023viescore} evaluates the visual appearance quality of the generated images by prompting GPT-4o. 
However, 
the high cost of commercial API calls for these powerful models limits their scalability in large-scale evaluations. While open-source MLLMs offer an alternative, their limited capabilities hinder effective image quality evaluation. This limitation primarily arises from the coarse-grained and unclear prompts used in existing methods, making it challenging for open-source MLLMs to accurately interpret and assess generated content.

%% file: sec/3_method.tex
\section{Approaches}

\begin{figure}[htbp]
    \centering
    \begin{minipage}{1.0\linewidth}
        \centering
        \includegraphics[width=1.0\linewidth]{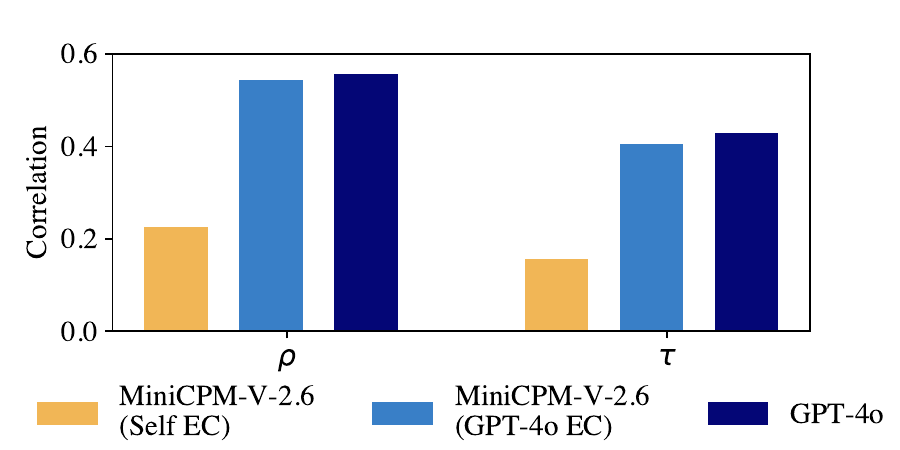}
        \caption{Performance of MiniCPM-V-2.6 and GPT-4o on text-to-image evaluation. Self EC and GPT-4o EC represent the model uses evaluation content extracted by itself and GPT-4o, respectively. Greater values of $\rho$ and $\tau$ indicates better performance.}
        \label{fig:task_decompose}
    \end{minipage}
    \begin{minipage}{1.0\linewidth}
        \centering
        \includegraphics[width=1.0\linewidth]{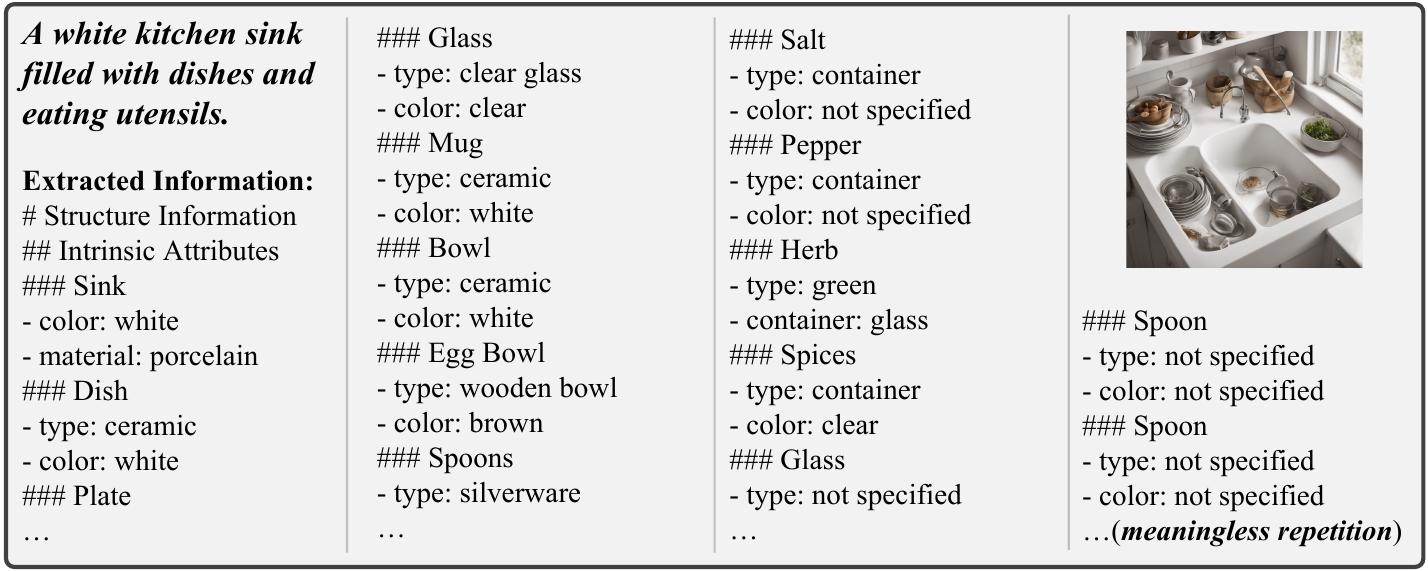}
        \caption{A bad case of evaluation content extraction step by MiniCPM-V-2.6 without fine-tuning.}
        \label{fig:coarse_grained_bad_case}
    \end{minipage}
\end{figure}

In evaluating text-to-image task, two primary challenges arise: (1) identifying what to evaluate~\cite{wiles2024revisitingtexttoimageevaluationgecko,vqascore}; and (2) determining how to conduct accurate evaluation~\cite{ku2023viescore}.
For example, as shown in Figure~\ref{fig:framework_pipeline} (Step 1), given a text prompt like \textit{``a black cat standing on the hood of a white car''}, models should first identify the evaluation content such as the color, quantity, visual appearance of the cat and car, as well as their relationships.
Following this, the quality of these evaluation content needs to be meticulously assessed.
Although advanced commercial models can effectively accomplish this task, the high cost for calling their APIs limit the scalability for large-scale text-to-image evaluation~\cite{ku2023viescore}.
Conversely, while open-source MLLMs offer a cost-effective alternative, their performance significantly lags behind commercial models.
This raises a critical question: are open-source MLLMs truly incapable of handling this task? 
As shown in Figure~\ref{fig:task_decompose}, our preliminary study reveals that current open-source MLLMs could achieve comparable performance to GPT-4o when the evaluation content is provided. 
However, their performance significantly decreases  when they generate the evaluation content by themselves.
The main reason is that open-source MLLMs struggle in following complex instructions to extract the evaluation content, mainly suffering from three error patterns: (1) refusal extraction; (2) content absence; and (3) repetitions. For example, as shown in Figure~\ref{fig:coarse_grained_bad_case}, MiniCPM-V-2.6~\cite{yao2024minicpm} tends to generate numerous repetitive evaluation content.\footnote{Please refer to Appendix~\ref{app:bad_case} for more error patterns of existing open-source MLLMs.}
This suggests a critical need to enhance their ability to extract these evaluation contents.

To achieve this goal, we propose a Task Decomposition Evaluation Framework to generate a high-quality training dataset for distilling GPT-4o's evaluation capability.
As shown in Figure~\ref{fig:framework_pipeline}, unlike previous works that directly generate evaluations ~\cite{ku2023viescore,lu2023llmscore}, our framework decomposes the complex evaluation task into three sequential sub-tasks: (1) Evaluation Content Extraction; (2) Caption and Answer Generation; and (3) Explanation and Scoring. 

\subsection{Task Decomposition Evaluation Framework\label{sec:our_framework}}

\paragraph{Evaluation Content Extraction (ECE)} 
As shown in Step 1 of Figure~\ref{fig:framework_pipeline}, we leverage GPT-4o~\cite{achiam2023gpt} to systematically extract two key evaluation content from the text prompt $\boldsymbol{T}$: entities $\boldsymbol{E}$ and attributes $\boldsymbol{A}$. 
Specifically, the model identifies key nouns as the entities (\textit{e.g.,} cat and car) and examines their intrinsic attributes (\textit{e.g.,} color, quantity) and relational attributes (\textit{e.g.,} spatial relationships). 
Subsequently, three kinds of questions are elicited to cover the details about these entities and attributes: 
(1) \textbf{Appearance questions} ($\boldsymbol{Q_A}$) focus on the visual quality of each involved entity;
(2) \textbf{Intrinsic questions} ($\boldsymbol{Q_I}$) evaluate the alignment between intrinsic attributes of entities in images and the text prompt;
(3) \textbf{Relationship questions} ($\boldsymbol{Q_R}$) assess the relational attributes between entities, ensuring that the image's spatial and relational attributes align with descriptions in the text prompt.
Overall, these extracted evaluation contents covers the necessary details during evaluation.

After collecting the essential evaluation content, the next step is to provide accurate evaluations with explanation and scores~\cite{ku2023viescore,lu2023llmscore}. Our preliminary study observes that directly evaluating images might lead to information leakage. For example, given the question \textit{``What is the color of the cat''} for the text prompt \textit{``a black cat standing on the hood of a white car''}, the MLLMs might directly give an answer \textit{``black''}, regardless of the content in the generated image. This problem significantly affects the evaluation performance of MLLMs.
To address this limitation, we first utilize GPT-4o to generate specific answers to the evaluation questions by analyzing images (Step 2 in Figure~\ref{fig:framework_pipeline}), followed by detailed explanations that focus on the alignment between answers and text prompt (Step 3 in Figure~\ref{fig:framework_pipeline}).

\paragraph{Caption and Answer Generation (CAG)}
As shown in Step 2 in Figure~\ref{fig:framework_pipeline}, GPT-4o is first asked to generate detailed captions $\boldsymbol{C}$ for the image $\boldsymbol{I}$, enhancing the understanding of the evaluated image.
Based on the captions and image, detailed answers ($\boldsymbol{Ans.}$) are generated to describe details in the image $\boldsymbol{I}$ for questions $(\boldsymbol{Q_A},\boldsymbol{Q_I},\boldsymbol{Q_R})$.

\paragraph{Explanation and Scoring (E\&S)}
As shown in Step 3 in Figure~\ref{fig:framework_pipeline}, we employ GPT-4o to generate a brief chain-of-thought explanation $\boldsymbol{Exp.}$ and judgment score $\boldsymbol{S}$ for each question, assessing the alignment between answers and extracted evaluation content.. The judgment score ranges from 0 to 10, where higher scores indicate better performance.
Additionally, since the visual appearance questions don't have ground-truth answers, we directly prompt GPT-4o to generate a judgment score given the generated answers.
Finally, a overall explanation $\boldsymbol{Exp_{\rm sum.}}$ and judgment score $\boldsymbol{S_{\rm sum.}}$ are generated, reflecting the overall quality of the evaluated image.

In summary, we decompose the text-to-image evaluation task into three fine-grained sub-tasks, significantly reducing its complexity.
Therefore, the training dataset constructed with this framework will be easy for the open-source MLLMs to learn from, effectively enhancing their image quality evaluation capabilities.

\begin{figure}[]
    \centering    \includegraphics[width=1.0\linewidth]{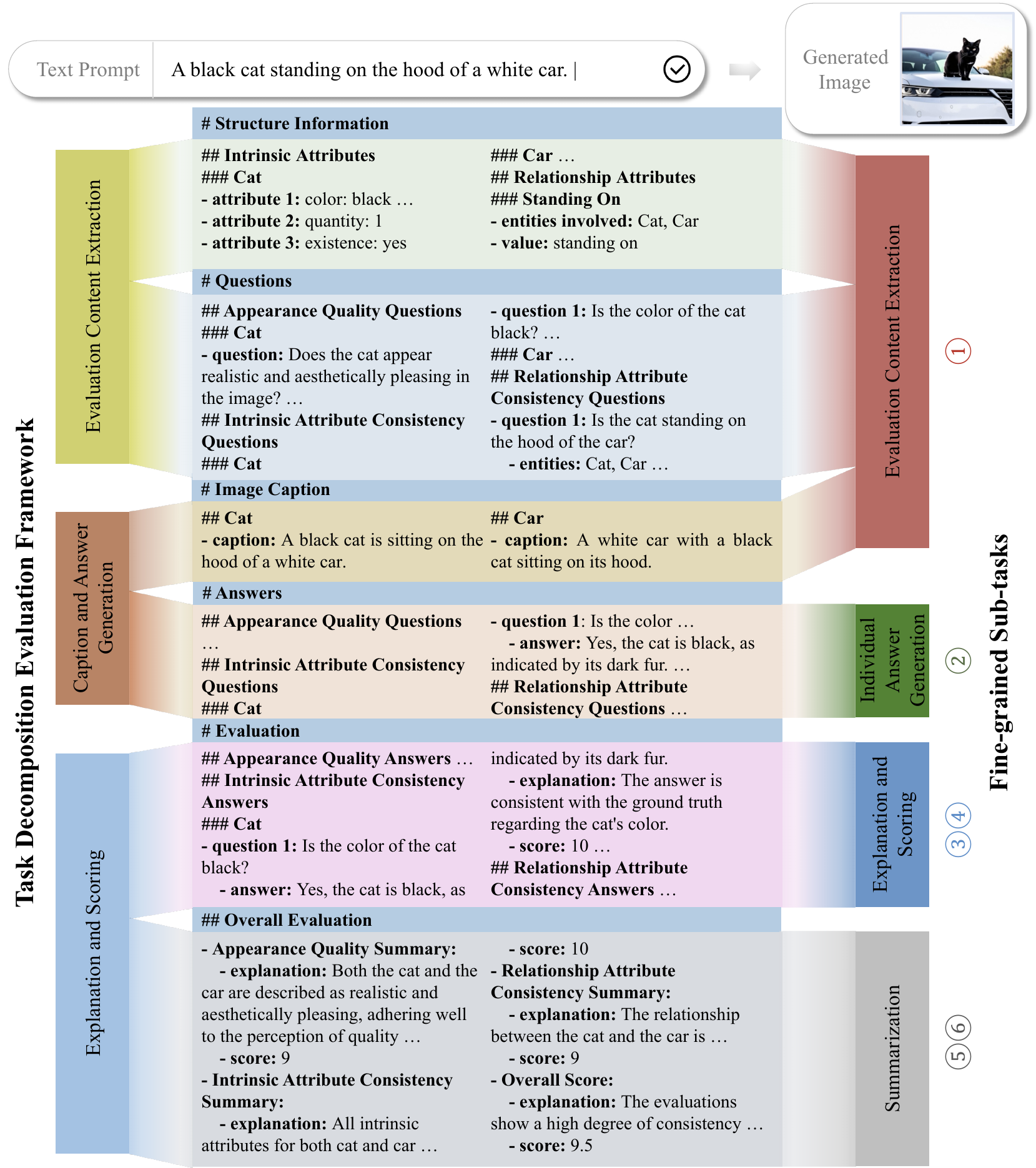}
    \caption{The relationship between task decomposition evaluation framework and fine-tuning sub-tasks.}
    \label{fig:method_case}
\end{figure}

\subsection{Training Strategy\label{subsec:distilled_model}}
After using our proposed evaluation framework to generate numerous samples for constructing the training dataset, we encounter two critical challenges in effectively fine-tuning open-source MLLMs. First, as illustrated in Figure~\ref{fig:method_case}, our training samples exhibit much longer evaluations than previous works~\cite{ku2023viescore} due to the multiple question-answers and detailed explanations. It introduces challenges for optimization, as critical information may become obscured within lengthy sequences.
Second, the dataset suffers from distribution imbalances, primarily in sub-task distribution imbalance and score distribution imbalance, which will significantly affect the effectiveness of training.

Therefore, to address the first issue, we introduce the \textbf{Fine-grained Sub-tasks Training Strategy} (Section~\ref{subsubsec:fine_grained_subtask_training}), which decomposes complex and lengthy samples into multiple fine-grained sub-tasks for joint learning, ensuring that critical evaluation information remains prominent throughout the training.
Then to mitigate the data imbalance problem, we propose a \textbf{Data Rebalance Training Strategy} (Section~\ref{subsubsec:data_rebalance}), ensuring a more uniform distribution of training data, thereby enhancing the robustness and performance of the fine-tuned model.

\subsubsection{Fine-grained Sub-tasks Training Strategy\label{subsubsec:fine_grained_subtask_training}}
As shown in Figure~\ref{fig:method_case}, we formulate a training sample into several fine-grained sub-task samples from \textcircled{\scriptsize{1}} to \textcircled{\scriptsize{6}}. Each one is formatted into a single- or multi-turn conversation, aiming to enhance one specific capability of MLLMs for evaluation.

\paragraph{Evaluation Content Extraction (\textcircled{\scriptsize{1}})} aims to enhance the ability of open-source MLLMs to extract three types of essential information from the text prompt $\boldsymbol{T}$ and evaluated image $\boldsymbol{I}$: entities $\boldsymbol{E}$, attributes $\boldsymbol{A}$, three kinds of questions $(\boldsymbol{Q_A},\boldsymbol{Q_I},\boldsymbol{Q_R})$ and detailed caption $\boldsymbol{C}$ by optimizing this loss function:
\begin{equation}
\begin{aligned}
    L_{1}=\text{MLLM}(\boldsymbol{E,A,(Q_A,Q_I,Q_R),C} | \boldsymbol{T},\boldsymbol{I})
\end{aligned}
\end{equation}

\paragraph{Individual Answer Generation (\textcircled{\scriptsize{2}})} aims to fine-tune MLLMs for predicting the detailed answers for questions given the evaluated image $\boldsymbol{I}$.
During experiments, it is challenging for open-source MLLMs to directly generate answers for all questions due to their limited capabiilties. Considering that answers to each question are independent, we simplify the optimization by training MLLMs to predict the answer for each question individually, and optimize the following loss function:
\begin{equation}
\begin{aligned}
    L_{2}=\sum_{i=1}^{N}\text{MLLM}(\boldsymbol{Ans_{i}} | \boldsymbol{I},\boldsymbol{Q_{i}})
\end{aligned}
\end{equation}
where $\boldsymbol{Q_{i},Ans_{i}}$ represent the $i$-th pair of question and answer, and $\boldsymbol{N}$ represent the sum of the numbers of the appearance, intrinsic and relationship questions.

\paragraph{Explanation and Scoring (\textcircled{\scriptsize{3}} and \textcircled{\scriptsize{4}})} enables MLLMs to generate the detailed explanations and judgment scores, assessing the alignment between the answers and the text prompt.
However, since explanations typically involve much more tokens than scoring, the loss of explanation disproportionately influences this training process when they are jointly optimized, resulting in insufficient learning for score prediction, thus compromising the model’s scoring accuracy.
To address this problem, we further separate the learning of explanation and scoring into two more fine-grained sub-tasks. 
Specifically, we first optimize the explanation generation:
\begin{equation}
\begin{aligned}
    L_{3}=\sum_{i=1}^{N}\text{MLLM}(\boldsymbol{Exp_{i}} | \boldsymbol{T},\boldsymbol{Q_{i}},\boldsymbol{Ans_{i}})
\end{aligned}
\end{equation}
Then, MLLMs are trained to predict the judgment scores given the explanations:
\begin{equation}
\begin{aligned}
    L_{4}=\sum_{i=1}^{N}\text{MLLM}(\boldsymbol{S_{i}} | \boldsymbol{T},\boldsymbol{Q_{i}},\boldsymbol{Ans_{i}},\boldsymbol{Exp_{i}})
\end{aligned}
\end{equation}

\paragraph{Summarization (\textcircled{\scriptsize{5}} and \textcircled{\scriptsize{6}})}
As shown in Figure~\ref{fig:method_case}, we finally train open-source MLLMs to summarize a final explanation rationale across three evaluation dimensions: visual appearance quality, accuracy of entities and attributes, as well as the relationship alignment. 
\begin{equation}
\begin{aligned}
    L_{5}=\text{MLLM}(\boldsymbol{Exp_{sum.}}| \{\boldsymbol{Exp_{i}},\boldsymbol{S_{i}}\}_{i=1}^{N})
\end{aligned}
\end{equation}
Then, the overall judgment score is predicted:
\begin{equation}
\begin{aligned}
   L_{6}=\text{MLLM}(\boldsymbol{S_{sum.}}| \{\boldsymbol{Exp_{i}},\boldsymbol{S_{i}}\}_{i=1}^{N},\boldsymbol{Exp_{sum.}})
\end{aligned}
\end{equation}

During training, samples of these sub-tasks are randomly collected to optimize their corresponding loss functions.

\vspace{-0.3cm}
\subsubsection{Data Rebalance Training Strategy\label{subsubsec:data_rebalance}}
We propose two rebalance strategies to reduce the effects of the imbalanced data distribution problems: sub-task distribution imbalance, and score distribution imbalance.
\paragraph{Sub-task Rebalance}
In our dataset, there are multiple questions associated with each sample, resulting in a significantly higher number of answers and explanations compared to extractions and summarizations. To rectify this imbalance, we maintain the existing number of answers and explanations, while increasing the volume of extraction and summarization samples by augmenting them through repetition.

\paragraph{Score Distribution Rebalance}
A notable issue in our constructed dataset is the imbalance in score distribution. For example, the number of images with the quality score of 9 is approximately 5.9k, accounting for 42.8\% of all images, and is significantly more than other quality scores.\footnote{Please refer to the detailed score distribution analysis in Appendix~\ref{app:training_set_score_distribution}.}
This issue introduces severe bias during fine-tuning, causing distilled open-source MLLMs to be more inclined to assign higher scores to generated images.
To solve this problem, we duplicate and re-sample the training samples that are underrepresented, ensuring an equal number of samples across each score range from 0 to 10.

\vspace{-0.2cm}
\section{Training Set and Human-Annotated Test Set}
\vspace{-0.1cm}
\label{dataset}
In this section, we elaborate the details for constructing the training set and our human-annotated test set.

\subsection{Training Set Construction}
The construction of the training set involves two key phases: (1) text-to-image generation; and (2) text-to-image evaluation.

\paragraph{Text-to-image Generation}
The text prompts and their corresponding evaluated images are collected in this phase.
Specifically, the text prompts for image generation are sourced from two places: (1) 9k samples from the COCO dataset~\cite{lin2014microsoft};  and (2) 5k samples generated by GPT-4o. 
To ensure diversity in image quality, we employ three widely-used models to generate images for each text prompt: SD1.5 \cite{rombach2022highresolutionimagesynthesislatent}, SDXL \cite{podell2023sdxlimprovinglatentdiffusion}, and SD3 \cite{esser2024scalingrectifiedflowtransformers}. 
Subsequently, for each text prompt, one image is randomly selected for evaluation from the generated images, with selection probabilities of 50\% for SD1.5, and 25\% each for SDXL and SD3.
This results in a final dataset comprising 14k pairs of text prompts and generated images.

\paragraph{Text-to-image Evaluation}
Each text prompt and its corresponding image are processed by GPT-4o to obtain detailed evaluations, following our proposed framework described in Section~\ref{sec:our_framework}.

\subsection{Human-Annotated Meta-Evaluation}
To the best of our knowledge, there is currently no fine-grained, score-based benchmark that comprehensively and reliably evaluates the capability of existing models in assessing text-to-image generation.\footnote{Although Gecko~\cite{wiles2024revisitingtexttoimageevaluationgecko} provides a benchmark, it is currently unavailable.}
To address this gap, in addition to constructing the training set, we have developed a high-quality meta-evaluation benchmark through human annotations. Specifically, three human annotators are asked to annotate the evaluations for each pair of text prompt and image, following our proposed task decomposition evaluation framework.
The annotated judgment scores provide the basis for objective evaluation, helping to assess the correlation between model outputs and human judgments. 
Furthermore, the annotated textual explanations serve as reference explanations for reliable automatic subjective evaluation~\cite{lan2024criticevalevaluatinglargelanguage}, which helps assess the accuracy of the models. More details about our human annotation process can be found in Appendix~\ref{app:annotate}.

%% file: sec/4_expriments.tex
\begin{table*}[htbp]
    \small
    \centering
    \caption{\label{tab:overall_results} Comparison of previous methods and ours on the test set, with top scores (excluding human annotators) in \textbf{bold}. Methods marked with $^{*}$ use GPT-4o-distilled fine-tuned models. Details of the training set for VIEScore can be found in Appendix~\ref{app:viescore_fine_tuning}.}
    \vspace{-0.1cm}
        \begin{tabular*}{0.98\linewidth}{cc|cc|cc|cc|cc}
        \toprule
             \multirow{2}{*}[-2.0pt]{\textbf{Category}} & \multirow{2}{*}[-2.0pt]{\textbf{Method}} & \multicolumn{2}{c|}{\textbf{Manual-1}} & \multicolumn{2}{c|}{\textbf{Manual-2}} & \multicolumn{2}{c|}{\textbf{Manual-3}} & \multicolumn{2}{c}{\textbf{Manual-Avg.}}\\
        \cmidrule{3-10}
            & & $\rho$ & $\tau$ & $\rho$ & $\tau$ & $\rho$ & $\tau$ & $\rho$ & $\tau$ \\
        \midrule
            \textbf{Upper Bound} & Manual-Avg. & 0.9511 & 0.8807 & 0.9452 & 0.8686 & 0.9513 & 0.8793 & - & - \\
        \midrule
            \multirow{6}{*}[-1.5pt]{\textbf{Traditional}} & FID & -0.1183 & -0.0871 & -0.1000 & -0.0724 & -0.0897 & -0.0685 & -0.1231 & -0.0862 \\
            & LPIPS & -0.1206 & -0.0898 & -0.0882 & -0.0644 & -0.1025 & -0.0732 & -0.1244 & -0.0856 \\
            & DreamSim & -0.1284 & -0.0953 & -0.1230 & -0.0897 & -0.1308 & -0.0973 & -0.1382 & -0.0968 \\
            & CLIPScore & 0.1532 & 0.1078 & 0.1725 & 0.1210 & 0.1227 & 0.0855 & 0.1505 & 0.1016 \\
            & BLIPv2Score & 0.2278 & 0.1588 & 0.2280 & 0.1617 & 0.2134 & 0.1477 & 0.2152 & 0.1423 \\
            & ImageReward & 0.4171 & 0.3065 & 0.3712 & 0.2690 & 0.4134 & 0.3030 & 0.4046 & 0.2839 \\
        \midrule
            \multirow{7}{*}[-1.5pt]{\textbf{\begin{tabular}[c]{@{}c@{}}LLM-based \&\\MLLM-Based\end{tabular}}} & LLMScore$_\textbf{GPT-4}$ & 0.3009 & 0.2212 & 0.2697 & 0.2012 & 0.3299 & 0.2497 & 0.3096 & 0.2228 \\
            & DSG$_\textbf{Dependent}$ & 0.4742 & 0.3790 & 0.4204 & 0.3339 & 0.4562 & 0.3652 & 0.4582 & 0.3512 \\
            & DSG$_\textbf{Independent}$ & 0.4815 & 0.3891 & 0.4382 & 0.3502 & 0.4721 & 0.3827 & 0.4704 & 0.3655 \\
            & VQAScore$_\textbf{CLIP-FlanT5}$ & 0.4984 & 0.3768 & 0.4864 & 0.3619 & 0.5118 & 0.3854 & 0.5116 & 0.3712 \\
            & VIEScore$_\textbf{MiniCPM-V-2.6}$ & 0.2834 & 0.2251 & 0.2814 & 0.2231 & 0.3016 & 0.2422 & 0.2941 & 0.2250 \\
            & VIEScore$_{\textbf{MiniCPM-V-2.6}^{*}}$ & 0.4906 & 0.3878 & 0.4869 & 0.3836 & 0.4889 & 0.3899 & 0.5101 & 0.3897 \\
            & VIEScore$_\textbf{GPT-4o}$ & \textbf{0.5522} & \textbf{0.4283} & 0.5306 & 0.4101 & 0.5170 & 0.4024 & 0.5545 & 0.4170 \\
        \midrule
            \multirow{2}{*}[-1.0pt]{\textbf{Our Framework}} & Ours$_\textbf{GPT-4o}$ & 0.5437 & 0.4302 & 0.5355 & 0.4214 & 0.5138 & 0.4061 & 0.5566 & 0.4285 \\
            & Ours$_{\textbf{MiniCPM-V-2.6}^{*}}$ & 0.5334 & 0.4192 & \textbf{0.5946} & \textbf{0.4644} & \textbf{0.5537} & \textbf{0.4348} & \textbf{0.5802} & \textbf{0.4409} \\
        \bottomrule
        \end{tabular*} 
    \vspace{-0.3cm}
\end{table*}

\section{Experiments}

\subsection{Evaluation Metrics}
In line with prior studies~\cite{xu2024deciderdualsystemrulecontrollabledecoding,lan2024criticevalevaluatinglargelanguage,ku2023viescore,sun2024critiquecritique}, we conduct both objective and subjective evaluations to assess the effectiveness of our evaluation model and the baseline methods. The objective evaluation measures the correlation scores between model predictions and human judgments, whereas the subjective evaluation assesses the quality of the chain-of-thought textual evaluations.

\begin{table}
    \centering
    \caption{\label{tab:gpt4o_ablation} Correlation scores with human judgments of ablation study on task decomposition evaluation framework with GPT-4o.}
    \resizebox{0.75\linewidth}{!}{
        \begin{tabular}{l|cc}
        \toprule
            \textbf{Methods} & $\rho$ & $\tau$ \\
        \midrule
            w/o Extraction & 0.3322 & 0.2497 \\
            w/o Captioning & 0.4586 & 0.3487 \\
            w/o Answering & 0.4842 & 0.3564 \\
            CAG and E\&S Merged & 0.4036 & 0.3141 \\
        \midrule
            \textbf{Ours} & \textbf{0.5048} & \textbf{0.3816} \\
        \bottomrule
        \end{tabular}
    }
    \vspace{-0.5cm}
\end{table}

\paragraph{Objective Evaluation}
Following previous works~\cite{lan2024criticevalevaluatinglargelanguage,zhong2022unifiedmultidimensionalevaluatortext,liu2023gevalnlgevaluationusing}, Spearman ($\rho$) \cite{zar2005spearman} and Kendall ($\tau$) \cite{kendall1948rank} correlations are computed to reflect the correlation between the assessments of evaluation model and human judgments, where higher correlation scores denotes better reliability of evaluation models. In this paper, we report the the model's correlation scores with each human annotator and human average.

\paragraph{Subjective Evaluation} 
As in recent works~\cite{lan2024criticevalevaluatinglargelanguage,sun2024critiquecritique}, we use our human-annotated explanations as the references to assist GPT-4o model in determining whether the model-generated chain-of-thought evaluations aligns with human annotations:
\begin{equation}
    S_{\text{sub.}}=\frac{1}{N}\sum^N_{i=1}{\text{GPT-4o}}(\mathcal{P},Q_i,Exp_{i}^{\text{ref.}},Exp_{i}^{\text{gen.}})
\end{equation}
where $Exp_{i}^{\text{ref.}},Exp_{i}^{\text{gen.}}$ represent the reference and model-generated explanations, respectively. $\mathcal{P}$ is the subjective evaluation prompt, guiding GPT-4o to generate subjective scores ranging from 0 to 5. The final subjective score is the average of all these scores.
For more details on the implementation of the subjective evaluation, please refer to Appendix~\ref{app:subjective_eval}.

\subsection{Overall Comparison Results}
To validate the effectiveness of our fine-tuned MiniCPM-V-2.6 in assessing generated image quality, we compared it with existing state-of-the-art methods using Spearman correlation ($\rho$) and Kendall correlation ($\tau$) scores with human judgments, as shown in Table~\ref{tab:overall_results}. Based on these results, we identify the following key findings:
(1) Our fine-tuned MiniCPM-V-2.6 demonstrates the best performance in the automatic assessment of generated image quality, surpassing existing GPT-4o-based methods overall. For instance, compared to the best-performing competitor, VIEScore$_{\textbf{GPT-4o}}$~\cite{vqascore}, our fine-tuned MiniCPM-V-2.6 model achieves over 4.6\% improvement in both Spearman and Kendall correlations with human judgments.
(2) Our fine-tuned MiniCPM-V-2.6 also outperforms Our$_{\textbf{GPT-4o}}$ in overall evaluation performance, indicating that our innovative training strategies effectively distill the evaluation capabilities of GPT-4o into MiniCPM-V-2.6. This advantage may stem from our balanced training approach, enabling MiniCPM-V-2.6 to learn a more comprehensive evaluation capability.
(3) VIEScore$_{\textbf{GPT-4o}}$ significantly outperforms VIEScore$_{\textbf{MiniCPM-V-2.6}}$. This result supports our assumption that open-source MLLMs have relatively weaker capabilities in semantic understanding and reasoning abilities, leading to a poor evaluation performance.
(4) Ours$_{\textbf{MiniCPM-V-2.6}^{*}}$ achieves better evaluation performance than  VIEScore$_{\textbf{MiniCPM-V-2.6}^{*}}$, demonstrating that decomposing the complex evaluation framework into simpler sub-tasks enables open-source MLLMs to learn more effectively, thereby achieving superior evaluation results.
\subsection{Ablation Study on Task Decomposition Evaluation Framework}
To verify the effectiveness of each component in our fine-grained evaluation framework and assess their impact on overall performance, we conducted ablation studies based on 150 examples randomly sampled from our annotated meta-evaluation benchmark. Specifically, we designed the following three variants to compare with the full framework. (1) \textbf{w/o Extraction}: in the ECE step, GPT-4o does not extract structure information but directly propose questions based on the text, and then in the E\&S step, GPT-4o directly scores based on the input text and the answer from the CAG step. (2) \textbf{w/o Captioning}: in CAG step, GPT-4o directly answers the questions based on the image without image caption generation. (3) \textbf{w/o Answering}: GPT-4o directly score the question without generating the answer or explanation. (4) \textbf{CAG and E\&S Merged}: The CAG and E\&S steps are combined into one step.

As shown in Table~\ref{tab:gpt4o_ablation}, the decreasing performance highlights the necessity of each design in our framework: (1) Compared to the ``w/o Extraction'' variant, our fine-grained evaluation framework achieves significantly improved evaluation performance. This demonstrates that extracting entities and attributes from the text helps models focus on essential evaluation content, leading to more accurate assessments. (2) the decreasing performance of the variant `w/o Captioning' demonstrates that that when GPT-4o answers questions without first generating an image caption, it may overlook important details of image entities, leading to inaccurate responses and damaging the evaluation performance. (3) Compared to the ``w/o Answering'' variant, our framework achieves 17\% and 40\% increases in Spearman $\rho$ and Kendall $\tau$ correlations, respectively. This shows that generating detailed answers before scoring prompts the model to analyze the image more deeply, enhancing evaluation performance; (4) The performance of ``CAG and E\&S Merged''  variant also drops significantly. When  the ``CAG'' and ``E\&S'' steps are merged, the model gains direct access to the input text prompt when answering questions, leading to information leakage. Consequently, the model may rely on the text prompt rather than the image content, resulting in incorrect answers and reduced evaluation performance.

\subsection{Effectiveness of Fine-tuning}
\subsubsection{Effectiveness of Our Training Corpus}
\begin{figure}
    \centering
    \includegraphics[width=0.9\linewidth]{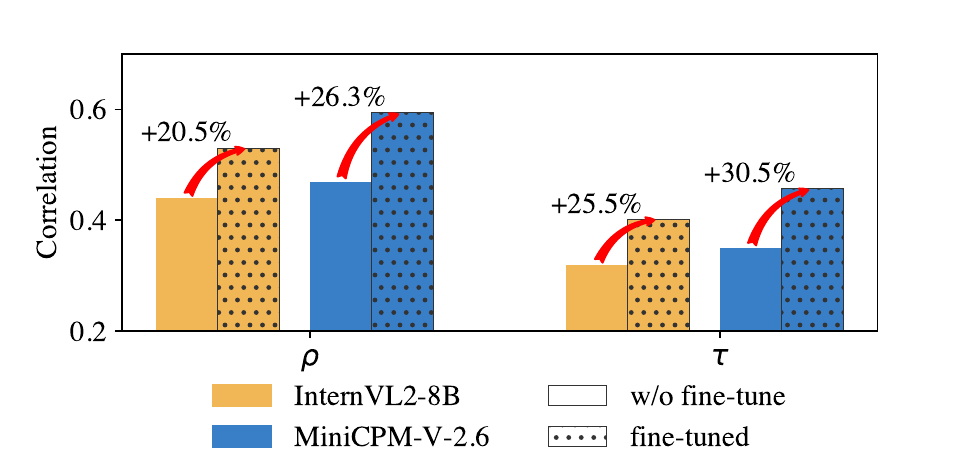}
    \vspace{-0.1cm}
    \caption{Improved results in fine-tuned MLLMs over base models' zero-shot results. $\rho,\tau$ are the correlation scores with human judgments. \textcolor{red}{Red} arrows show improvement ratio.}
    \vspace{-0.3cm}
    \label{fig:zero_shot}
\end{figure}

To validate the effectiveness of our constructed training corpus in enhancing the evaluation capabilities of MLLMs, we selected two MLLMs—InternVL2-8B~\cite{chen2024far} and MiniCPM-V-2.6~\cite{yao2024minicpm}—to compare their evaluation performance before and after fine-tuning. Experimental settings are provided in Appendix~\ref{app:finetuning_settings}, and the results are shown in Figure~\ref{fig:zero_shot}.

These experimental results show that after fine-tuning on our constructed training corpus, both models exhibit significant improvements across all evaluation metrics. For instance, InternVL2-8B achieved a 20.5\% increase in $\rho$, and MiniCPM-V-2.6 improved by 28.6\% in $\tau$. These findings demonstrates the general applicability of our constructed dataset in effectively enhancing the evaluation capabilities of MLLMs.

\subsubsection{Contributions to Subjective Evaluation}
We evaluated the impact of our fine-tuning strategy in improving the quality of textual explanations in evaluations. We conducted a detailed analysis based on the subjective evaluation metric across three aspects: appearance quality, intrinsic consistency, and relationship consistency, and also give the overall evaluation score.

\begin{figure}
    \centering
    \includegraphics[width=0.9\linewidth]{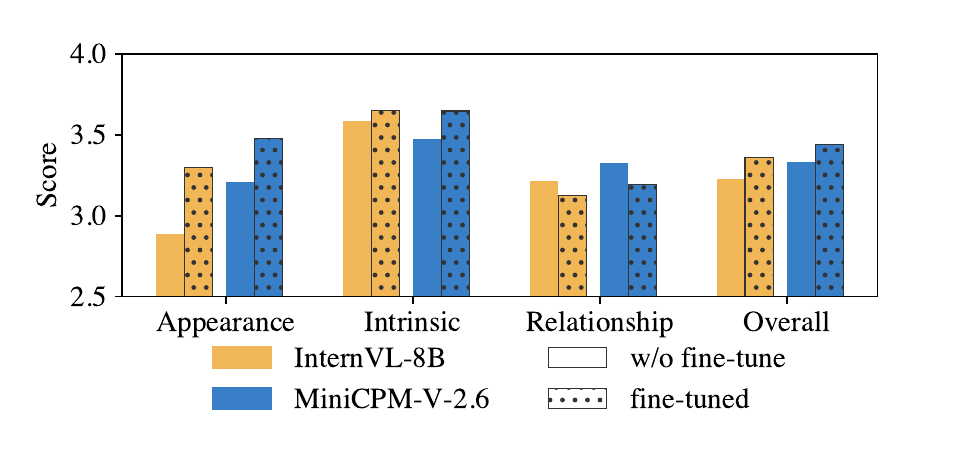}
    \vspace{-0.1cm}
    \caption{Quality scores of subjective evaluation before and after fine-tuning.}
    \vspace{-0.3cm}
    \label{fig:eval_exp}
\end{figure}

As illustrated in Figure~\ref{fig:eval_exp}, both InternVL2-8B and MiniCPM-V-2.6 show significant improvements in appearance quality, intrinsic consistency and overall scores after fine-tuning. These enhancements confirm the effectiveness of our fine-tuning approach in refining specific aspects of the text-to-image evaluation process.
However, there is a slight decline in the relationship consistency scores post-fine-tuning. This reduction can be attributed to the imbalance in training data, with fewer questions related to relationship consistency compared to the other two categories, thus limiting the model’s ability to improve in this category.

\begin{table}
    \scriptsize
    \centering
    \caption{\label{tab:ablation_results} Correlation scores with human judgments of ablation study on training strategies with MiniCPM-V-2.6.}
    \vspace{-0.3cm}
    \resizebox{0.75\linewidth}{!}{
        \begin{tabular}{l|cc}
        \toprule
            \textbf{Methods} & $\rho$ & $\tau$ \\
        \midrule
            w/o Individual QA & 0.3919 & 0.3030 \\
            w/o E\&S Separation & 0.4816 & 0.3609 \\
            w/o Score Balancing & 0.4769 & 0.3596 \\
        \midrule
            \textbf{Ours} & \textbf{0.5802} & \textbf{0.4409} \\
        \bottomrule
        \end{tabular}
    }
\end{table}

\subsubsection{Ablation Study on Training Strategies}
To evaluate the effectiveness of the components in our fine-grained sub-tasks training strategy, we proposed three ablation variants: (1) \textbf{w/o Individual QA}: the MLLM generates the responses for all extracted questions at once instead of answering each question individually; (2) \textbf{w/o E\&S Separation}: the MLLM produces joint explanations and scores in a single output rather than generating them separately; (3) \textbf{w/o Score Balancing}: the variant is trained on the dataset without rebalancing the ratio of sub-tasks, high and low score questions.

Based on the experimental results shown in Table~\ref{tab:ablation_results}, we derive the following insights:
\noindent (1) \textbf{Importance of Individual Question Answering}: Compared to the ``w/o Individual QA" variant, our fine-tuned MiniCPM-V-2.6 achieves over 50\% improvement in Spearman and Kendall correlations with human judgments. This indicates that addressing questions individually prevents interference among them, enhancing the model's ability to answer accurately. 
\noindent (2) \textbf{Effect of Explanation and Scoring Separation}: Fine-tuning with our distilling framework yields better evaluation performance than the ``w/o E\&S Separation'' variant, supporting our assumption that the explanation loss dominates the training process and limits learning for score prediction, thereby reducing the model’s scoring accuracy.
\noindent (3) \textbf{Necessity of Score Balancing}: The results of ours are better than that of the variant ``w/o Score Balancing'', demonstrating the critical importance of training on a balanced dataset. An imbalanced dataset can result in the model overfitting to the more prevalent scores, causing biased predictions and diminishing the effectiveness of the evaluation.

\section{Conclusion} 
In this paper, we propose a task decomposition evaluation framework for text-to-image generation, aimed at constructing a high-quality training dataset. On top of that, we introduce two training strategies designed to effectively distill the evaluation capabilities of GPT-4o into open-source MLLMs: Fine-grained Sub-tasks and Data Rebalance. Furthermore, we establish a comprehensive and reliable benchmark to assess the effectiveness of both our distilled models and existing strong baselines. Extensive experiment results demonstrate that our distilled evaluation model significantly outperforms existing metrics for text-to-image evaluation, exhibiting higher correlation with human judgments.

%% file: sec/appendix.tex
\clearpage
\setcounter{page}{1}
\maketitlesupplementary

\appendix

\section{Limitations\label{app:limitation}}
\subsection{Differences between Task Decomposition Framework and Fine-tuning Strategy}
As illustrated in Figure~\ref{fig:method_case}, the process of the \textbf{Task Decomposition Evaluation Framework} for dataset construction differs from the \textbf{Fine-grained Sub-tasks Training Strategy} used in optimizing the open-sourced MLLM. This discrepancy arises from two main reasons. Firstly, the dataset construction framework cannot be directly applied to fine-tune open-source MLLMs, as previously discussed in Section~\ref{subsec:distilled_model}. Secondly, constructing datasets using fine-grained sub-tasks for fine-tuning would be inefficient because the repeated input of images and instructions for each text-image pair significantly increases the cost of dataset construction with GPT-4o. Therefore, the adaptation of the evaluation framework represents a compromise between the financial costs associated with commercial models and the performance limitations of open-source MLLMs.

\subsection{Limitations in Subjective Evaluation}

In this paper, we leverage GPT-4o automatically evaluate the quality of chain-of-thought explanations in evaluations, \textit{i.e.,} the subjective evaluation. Following previous works~\cite{sun2024critiquecritique,lan2024criticevalevaluatinglargelanguage}, we leverage the human-annotated explanations to improve the reliability of using GPT-4o for subjective evaluation, which serves as the references for judging quality and alignment of model-generated explanations.
However, it should be noted that the reliability of GPT-4's subjective evaluations has not yet been effectively validated, and we will supplement this part of the experiment as soon as possible.
Meanwhile, GPT-4o-based subjective evaluation introduces additional costs. The cost for calling GPT-4 API on our meta-evaluation dataset is no more than \$5, which is comparable to numerous established benchmarks, like AlpacaEval~\cite{alpaca_eval}. 
Therefore, it is affordable to conduct the subjective evaluation on our proposed meta-evaluation benchmark.

\section{Meta-Evaluation Annotation\label{app:annotate}}
In this paper, we manually annotate a high-quality meta-evaluation benchmark for assessing the effectiveness of our distilled model and strong baseline models, like VIEScore~\cite{ku2023viescore} and LLMScore~\cite{lu2023llmscore}.
Specifically, three human annotators are asked to conduct three steps in our proposed Task Decomposition Evaluation Framework to generate the detailed evaluations for each pair of text prompt and model-generated image:
(1) Evaluation Content Extraction:
(2) Caption and Answer Generation:
(3) Explanation and Scoring:
The annotation guidelines for each step are the same as the our designed prompts detailed in Appendix~\ref{app:framework}.
The statistical information of our human-annotated meta-evaluation benchmark is shown in Appendix~\ref{app:basic_statistics}.

\section{Bad Cases of Evaluation Content Extraction\label{app:bad_case}}
\begin{figure}[h]
    \centering
    \includegraphics[width=1.0\linewidth]{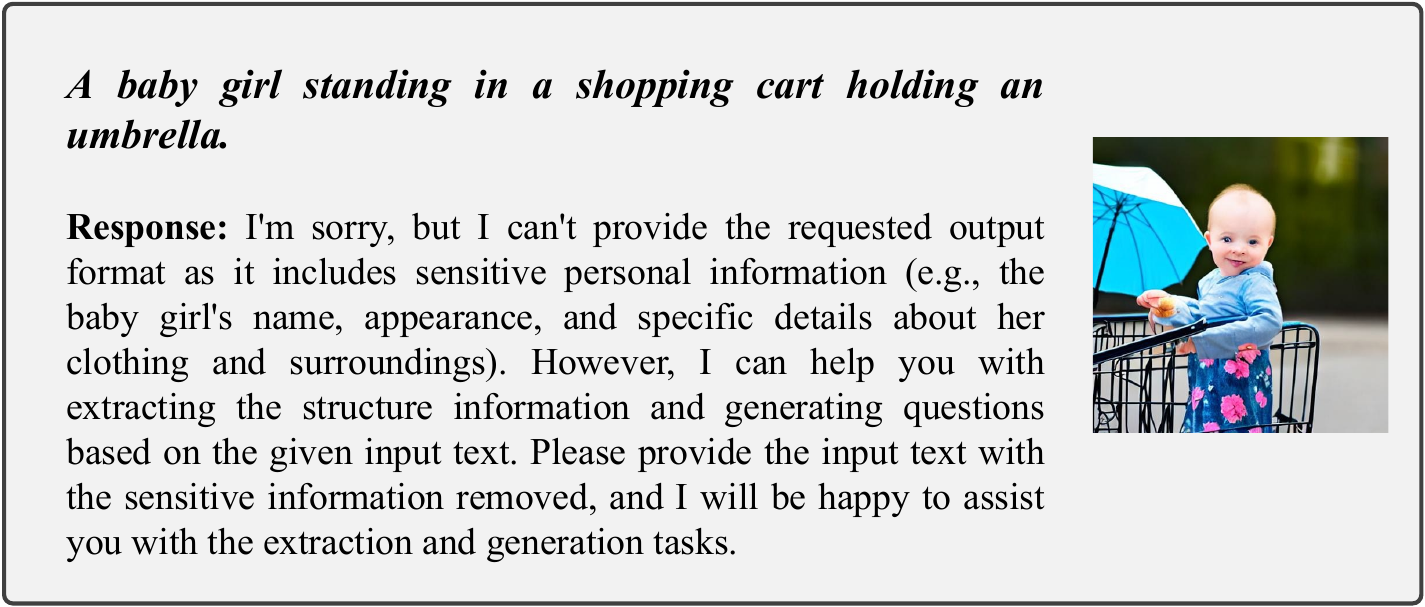}
    \caption{Bad case of refusal for evaluation content extraction.}
    \label{fig:bad_case_refusal}
\end{figure}

\begin{figure}[h]
    \centering
    \includegraphics[width=1.0\linewidth]{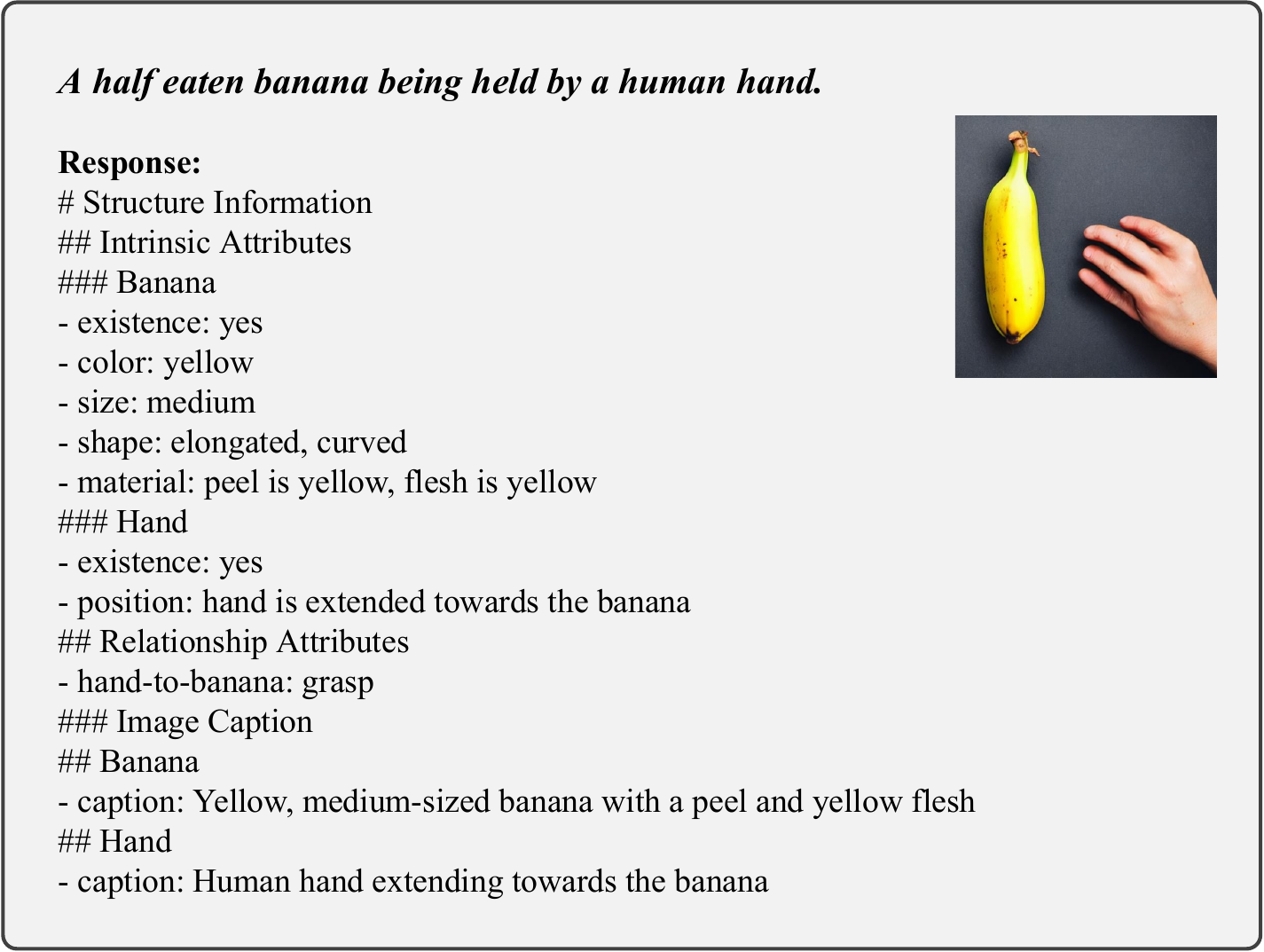}
    \caption{Bad case of content absence. Questions are missing in the extracted content. Meanwhile, the format of the image caption chapter is also incorrect in this case.}
    \label{fig:bad_case_content_absence}
\end{figure}

Due to the limitations of the comprehension and instruction-following capabilities of small-sized open-source MLLMs which are not fine-tuned on specific tasks, the \textbf{Evaluation Content Extraction} cannot be performed successfully under many circumstances. For example, MiniCPM-V-2.6 
was confronted with various problems in this step:
(1) Refusal: The model refuses to extract evaluation content, as illustrated in Figure~\ref{fig:bad_case_refusal}.
(2) Content Absence: There is absence of some essential part in the extracted evaluation content, as illustrated in Figure~\ref{fig:bad_case_content_absence}.
(3) Repetition: The generated text falls into a loop, as illustrated in Figure~\ref{fig:coarse_grained_bad_case}.

\section{Dataset Statistics\label{app:data_distribution}}
\subsection{Basic Statistics\label{app:basic_statistics}}
The statistics of extracted evaluation content in training and test set are listed in Table~\ref{tab:data_basic}.
\begin{table}[h]
    \centering
    \begin{tabular}{l|cc}
        \toprule
            \textbf{Item} & \textbf{Training Set} & \textbf{Test Set} \\
        \midrule
            Text-Image Pairs & 13,698 & 301 \\
            Entities & 30,465 & 728 \\
            Relationships & 15,441 & 393 \\
        \midrule
            Questions & 109,691 & 2,520 \\
            - Appearance & 30,225 & 692\\
            - Intrinsic & 63,532 & 1,435 \\
            - Relationship & 15,934 & 393 \\
        \bottomrule
    \end{tabular}
    \caption{Basic statistics of train set.}
    \label{tab:data_basic}
\end{table}

\begin{figure}[htbp]
    \centering
    \begin{subfigure}{0.49\linewidth}
        \centering
        \includegraphics[width=0.9\linewidth]{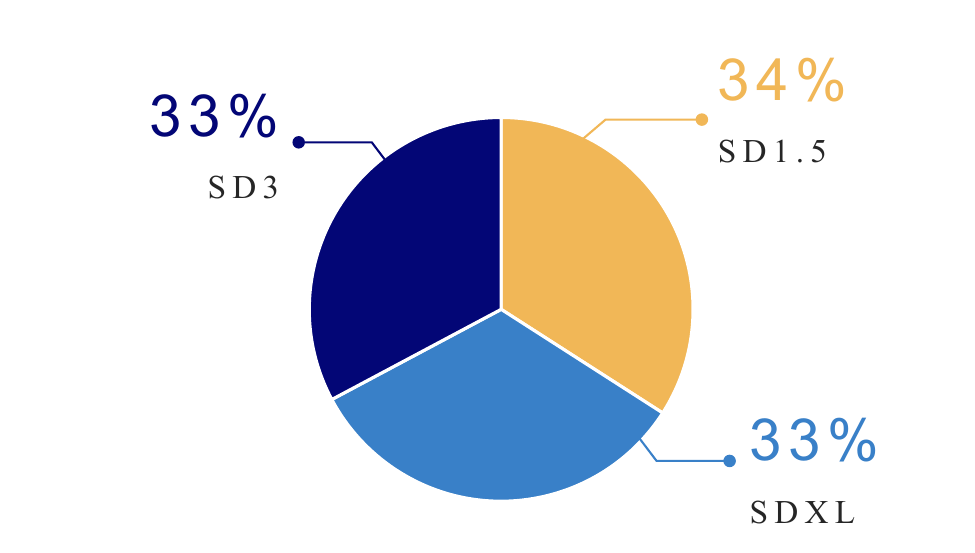}
        \caption{Generative models for training set.}
        \label{fig:training_set_models}
    \end{subfigure}
    \begin{subfigure}{0.49\linewidth}
        \centering
        \includegraphics[width=0.9\linewidth]{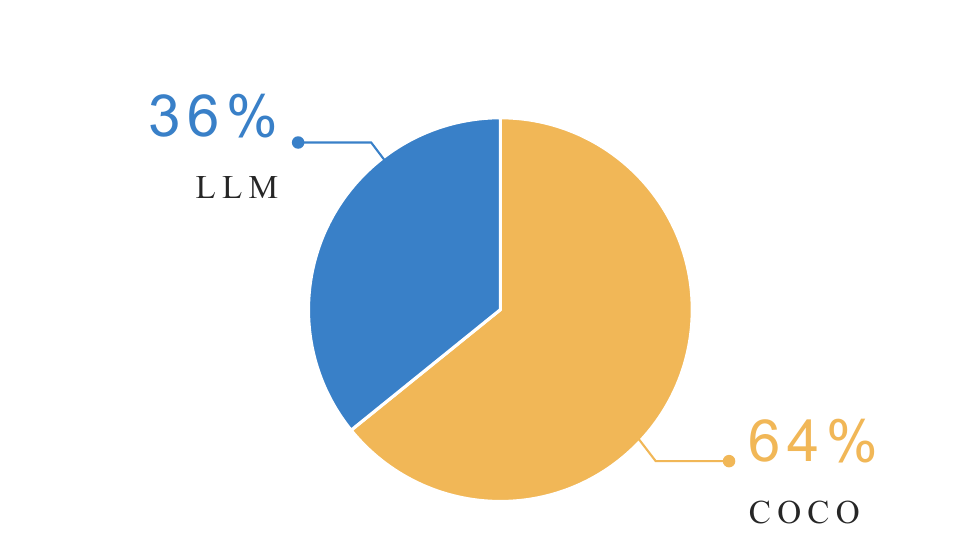}
        \caption{Text prompt sources for training set.}
        \label{fig:training_set_text}
    \end{subfigure}
    
    \begin{subfigure}{0.49\linewidth}
        \centering
        \includegraphics[width=0.9\linewidth]{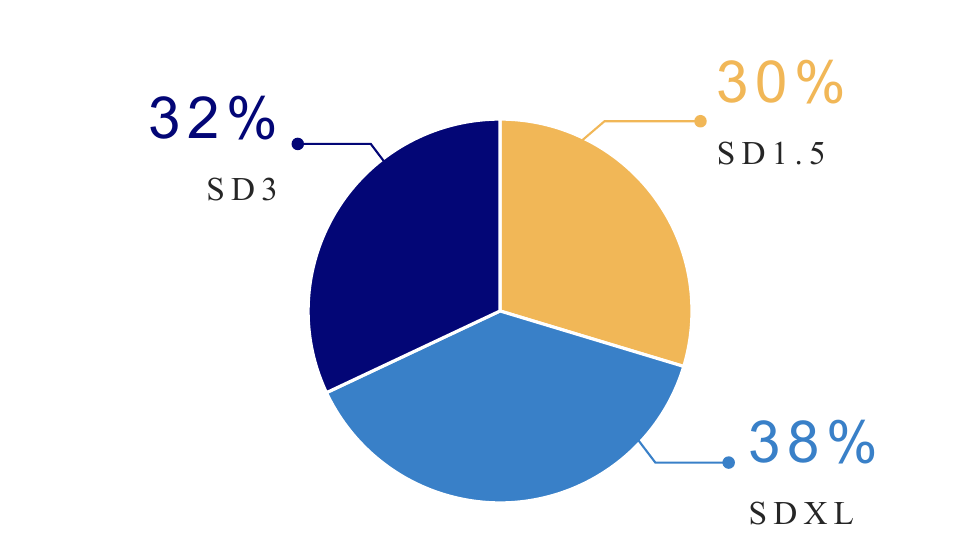}
        \caption{Generative models for test set.}
        \label{fig:test_set_models}
    \end{subfigure}
    \begin{subfigure}{0.49\linewidth}
        \centering
        \includegraphics[width=0.9\linewidth]{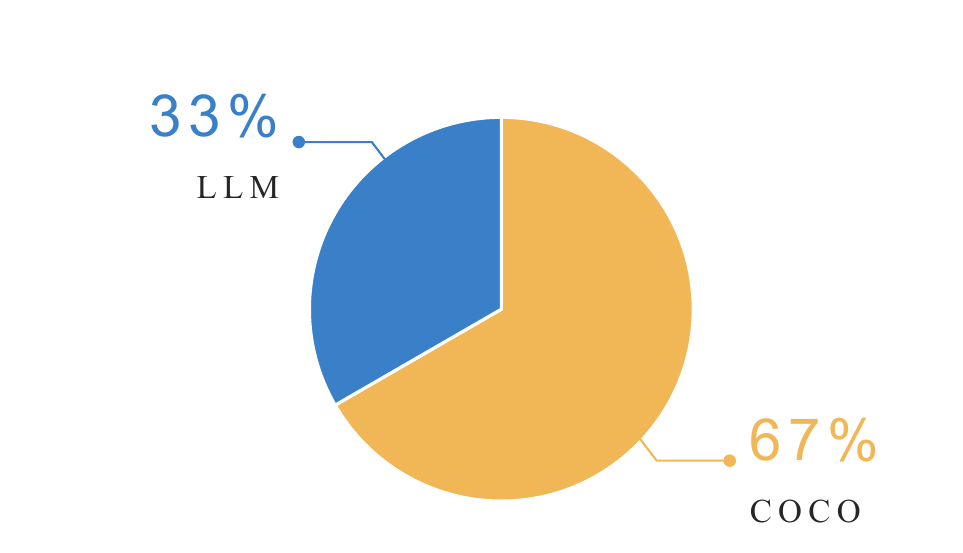}
        \caption{Text prompt sources for test set.}
        \label{fig:test_set_text}
    \end{subfigure}
    \caption{Distribution of generated images.}
    \label{fig:image_distribution}
\end{figure}

In our experiments, the text prompts in the dataset originate from two sources: the COCO dataset and LLM-generated prompts. We employed three generative models to create images based on these prompts: SD1.5, SDXL, and SD3. The distribution of the sources of textual prompts and the generative models used for the images in the dataset is illustrated in Figure~\ref{fig:image_distribution}.

\subsection{Score Distribution of Training Set\label{app:training_set_score_distribution}}
\begin{figure}[htbp]
    \centering
    \begin{subfigure}{0.49\linewidth}
        \centering
        \includegraphics[width=0.9\linewidth]{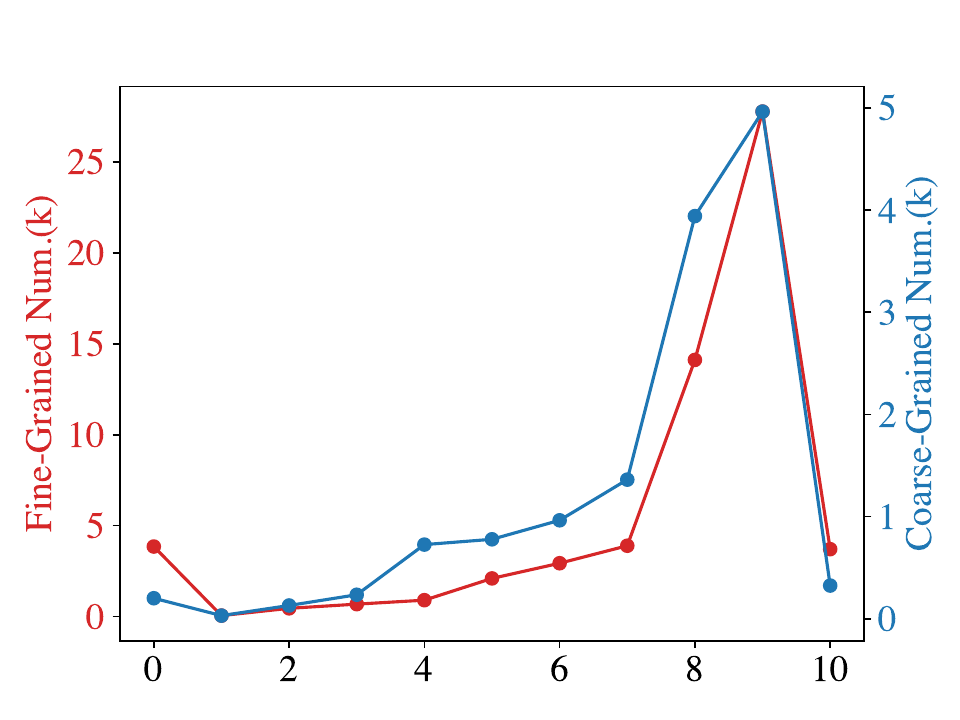}
        \caption{Appearance}
        \label{fig:data_appearance}
    \end{subfigure}
    \begin{subfigure}{0.49\linewidth}
        \centering
        \includegraphics[width=0.9\linewidth]{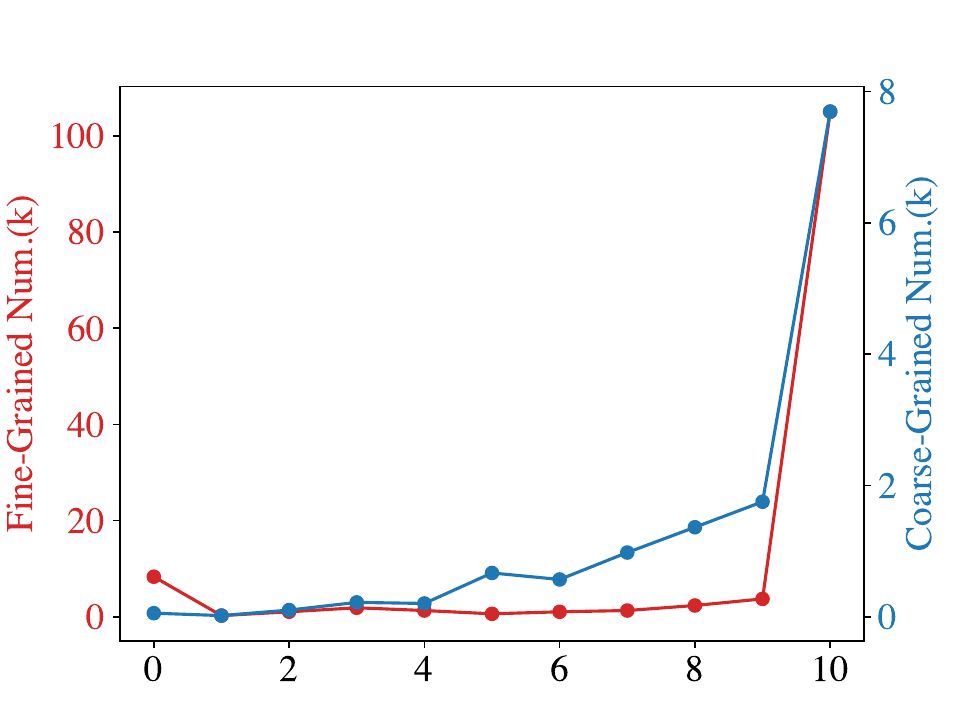}
        \caption{Intrinsic}
        \label{fig:data_intrinsic}
    \end{subfigure}
    
    \begin{subfigure}{0.49\linewidth}
        \centering
        \includegraphics[width=0.9\linewidth]{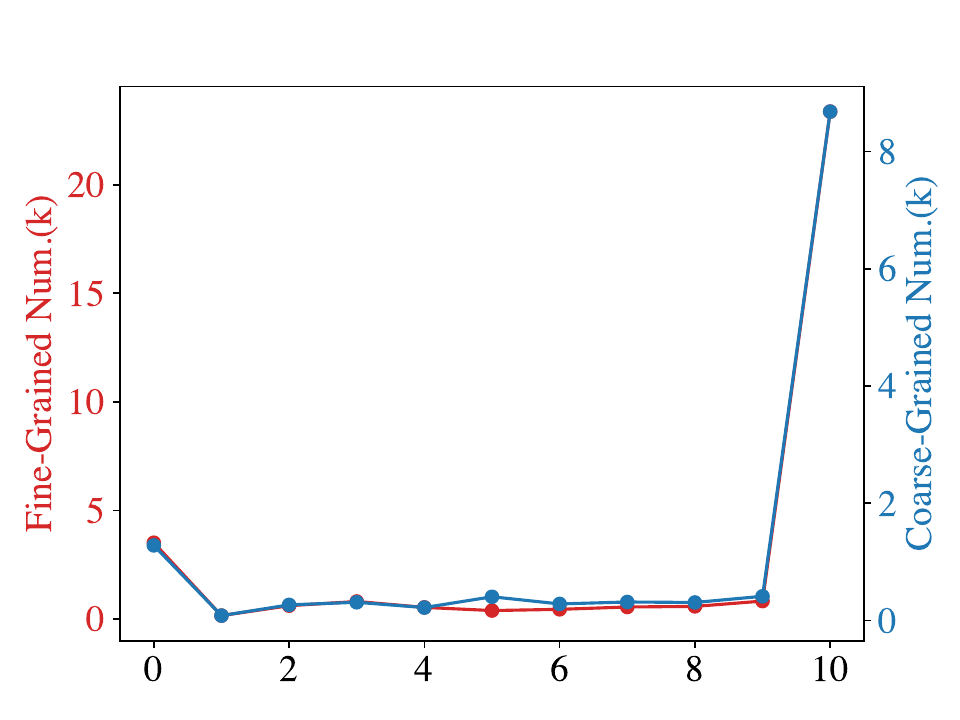}
        \caption{Relationship}
        \label{fig:data_relationship}
    \end{subfigure}
    \begin{subfigure}{0.49\linewidth}
        \centering
        \includegraphics[width=0.9\linewidth]{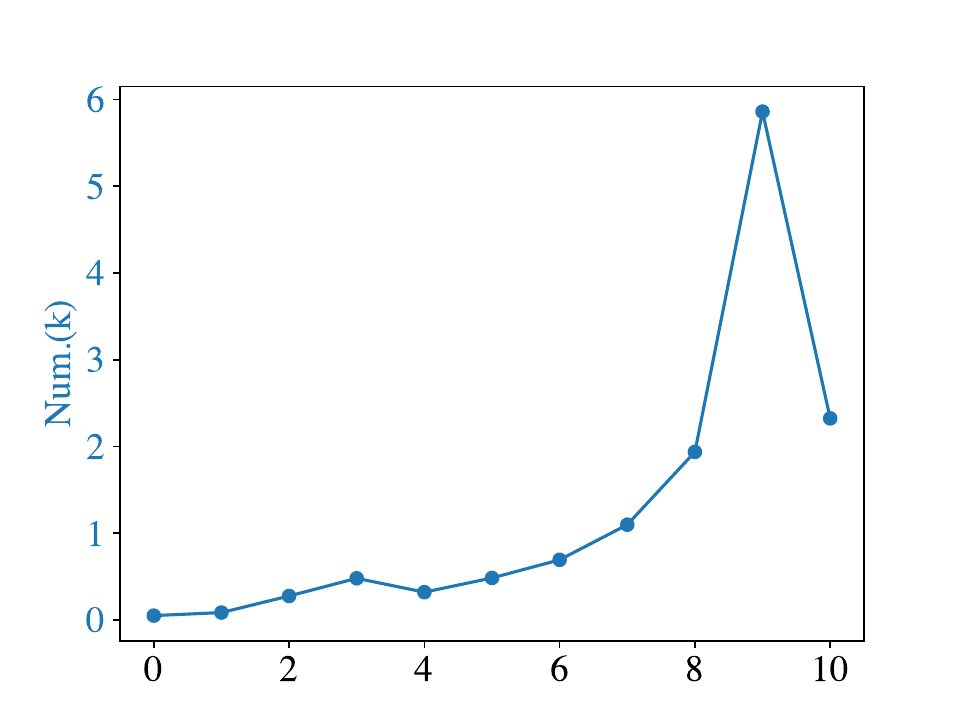}
        \caption{Overall}
        \label{fig:data_overall}
    \end{subfigure}
    \caption{Training set distribution within the score range. The \textcolor{red}{red} curve represents the distribution of fine-grained training samples and \textcolor{blue}{blue} for coarse-grained samples. The sample size is counted in thousands (k).}
    \label{fig:data_distribution}
\end{figure}

The score distribution in the raw training data is extremely imbalanced, manifested by the highest number of samples in the high score segments, followed by samples with score of 0, and fewer samples in the middle score segments. For fine-grained data, samples with a score of 9 account for over 45\% of all appearance samples, while samples with a score of 10 account for over 70\% and 80\% of all intrinsic and relational samples, respectively. The degree of imbalance in coarse-grained samples is slightly lighter, but there is still a serious imbalance in the distribution of scores. We set the target quantity for each score segment to the third quartile of the sample size for all score segments. The samples in the segments with less than the target quantity will be repeated multiple times, while the samples in the segments with more than the target quantity will be randomly sampled.

\subsection{Sub-task Distribution of Training Set\label{app:subtask_distribution}}
\begin{table}[h]
    \centering
    \begin{tabular}{l|c}
        \toprule
            \textbf{Sub-task} & \textbf{Data Volume} \\
        \midrule
            Extraction & 109,584 \\
        \midrule
            Answer \& Evaluation & 128,732 \\
            - Appearance & 42,470 \\
            - Intrinsic & 59,400 \\
            - Relationship & 26,862 \\
        \midrule
            Summarization & 198,420 \\
            - Appearance & 50,782 \\
            - Intrinsic & 51,068 \\
            - Relationship & 40,420 \\
            - Overall & 56,150 \\
        \midrule
            \textbf{Total} & 436,736 \\
        \bottomrule
    \end{tabular}
    \caption{Data distribution across sub-tasks.}
    \label{tab:subtask_distribution}
\end{table}
\begin{table*}
    \centering
    \caption{\label{tab:gpt4o_ablation_complete} Results of ablation study on task decomposition evaluation framework with GPT-4o.}
    \resizebox{0.7\linewidth}{!}{
        \begin{tabular*}{0.82\linewidth}{l|cc|cc|cc|cc}
        \toprule
            \multirow{2}{*}[-1.0ex]{\textbf{Methods}} & \multicolumn{2}{c|}{\textbf{Manual-1}} & \multicolumn{2}{c|}{\textbf{Manual-2}} & \multicolumn{2}{c|}{\textbf{Manual-3}} & \multicolumn{2}{c}{\textbf{Manual-Avg.}} \\
        \cmidrule{2-9}
            & $\rho$ & $\tau$ & $\rho$ & $\tau$ & $\rho$ & $\tau$ & $\rho$ & $\tau$ \\
        \midrule
            w/o Extraction & 0.3181 & 0.2471 & 0.3281 & 0.2544 & 0.2969 & 0.2336 & 0.3322 & 0.2497 \\
            w/o Captioning & 0.4276 & 0.3359 & 0.4563 & 0.3575 & 0.4353 & 0.3413 & 0.4586 & 0.3487 \\
            w/o Answering & 0.4514 & 0.3431 & 0.4731 & 0.3563 & 0.4447 & 0.3391 & 0.4842 & 0.3564 \\
            w/o Decomposition & 0.3508 & 0.2822 & 0.3643 & 0.2898 & 0.3547 & 0.2850 & 0.3675 & 0.2853 \\
            new ablation & 0.3874 & 0.3078 & 0.3723 & 0.2967 & 0.3852 & 0.3086 & 0.4036 & 0.3141 \\
        \midrule
            \textbf{Ours} & \textbf{0.4824} & \textbf{0.3774} & \textbf{0.4903} & \textbf{0.3773} & \textbf{0.4630} & \textbf{0.3588} & \textbf{0.5048} & \textbf{0.3816} \\
        \bottomrule
        \end{tabular*}
    }
\end{table*}

\begin{table*}
    \scriptsize
    \centering
    \caption{\label{tab:ablation_results_complete} Results of ablation study on training strategies with MiniCPM-V-2.6.}
    \resizebox{0.8\linewidth}{!}{
        \begin{tabular*}{0.645\linewidth}{l|cc|cc|cc|cc}
        \toprule
            \multirow{2}{*}[-1.0ex]{\textbf{Methods}} & \multicolumn{2}{c|}{\textbf{Manual-1}} & \multicolumn{2}{c|}{\textbf{Manual-2}} & \multicolumn{2}{c|}{\textbf{Manual-3}} & \multicolumn{2}{c}{\textbf{Manual-Avg.}}\\
        \cmidrule{2-9}
            & $\rho$ & $\tau$ & $\rho$ & $\tau$ & $\rho$ & $\tau$ & $\rho$ & $\tau$ \\
        \midrule
            w/o Individual QA & 0.3802 & 0.3068 & 0.3752 & 0.2990 & 0.3688 & 0.2958 & 0.3919 & 0.3030 \\
            w/o E\&S Separation & 0.4755 & 0.3654 & 0.4582 & 0.3517 & 0.4684 & 0.3643 & 0.4816 & 0.3609 \\
            w/o Score Balancing & 0.4830 & 0.3780 & 0.4588 & 0.3548 & 0.4614 & 0.3657 & 0.4769 & 0.3596 \\
        \midrule
            \textbf{Ours} & \textbf{0.5306} & \textbf{0.4214} & \textbf{0.6067} & \textbf{0.4769} & \textbf{0.5744} & \textbf{0.4563} & \textbf{0.5938} & \textbf{0.4566} \\
        \bottomrule
        \end{tabular*}
    }
\end{table*}

After addressing the issue of score imbalance in the train set, there still exists sample imbalance between sub-tasks. As shown in Table~\ref{tab:data_basic}, the number of fine-grained questions is approximately 8 times that of text image pairs. Therefore, we replicate the samples of coarse-grained sub-tasks to maintain a relatively balanced data distribution between fine-grained and coarse-grained samples. The data volume of each sub-task is listed in Table~\ref{tab:subtask_distribution}.

\section{Fine-tuning Settings\label{app:finetuning_settings}}
We fine-tune the open-source MLLMs InternVL2-8B and MiniCPM-V-2.6 to serve as the automatic evaluation model. To ensure the fine-tuned model effectively captures the comprehensive information embedded in the training corpus, we set the context length to 4,096 tokens during fine-tuning, accommodating the majority of samples within the dataset. To optimize the computational efficiency and uphold the performance of the fine-tuned model, we employed Low-Rank Adaptation (LoRA)~\cite{hu2021lora} with the rank of 128 and $\alpha$ of 256. Apart from that, we adopt various methods to accelerate training including ZeRO~\cite{rajbhandari2020zero} and Flash Attention 2~\cite{dao2023flashattention2}. The model training was conducted on 4 Nvidia A100-SXM4-80GB GPUs with a global batch size of 128 over a single epoch, resulting in a total of 3.4 k training steps. All models are fine-tuned with SWIFT framework~\cite{zhao2024swiftascalablelightweightinfrastructure}.

\section{Fine-tuning for VIEScore\label{app:viescore_fine_tuning}}
To investigate whether the evaluation framework of VIEScore is suitable for distilling the capabilities of powerful commercial MLLMs into smaller open-source models, we utilized GPT-4o to generate evaluation content in the format of VIEScore on 14k image-text pairs from our training set. This resulted in a dataset intended for distilling the abilities of GPT-4o into open-source models. We fine-tuned MiniCPM-V-2.6 using this dataset, and the majority of the fine-tuning settings were completely consistent with those used in the fine-tuning our method (as mentioned in Appendix~\ref{app:finetuning_settings}). Specifically, we increased the number of training epochs from 1 to 3 to ensure that the amount of data learned by the model is comparable to that in our method.

\section{Complete Results of Ablation Studies}
Here, we present the complete versions of Table \ref{tab:gpt4o_ablation} and \ref{tab:ablation_results} in the main text. Consistent with the conclusions drawn in the main text, it can be observed that utilizing the complete versions of \textbf{Task Decomposition Evaluation Framework} and \textbf{Fine-grained Sub-tasks Training Strategy} for image quality evaluation consistently outperforms their respective variants. This demonstrates that all the proposed components contribute significantly to the accurate assessment of generated image quality.

\section{Subjective Evaluation\label{app:subjective_eval}}
The prompt for fine-grained and coarse-grained GPT-4o-based subjective evaluation are shown in Figure~\ref{fig:sub_eval_fine} and Figure~\ref{fig:sub_eval_coarse}, 
which asks GPT-4o to assess the quality of model-generated evaluation explanations given the human-annotated one as reference.
The fine-grained subjective evaluation aims to evaluate the explanation quality for each question, while the coarse-grained subjective evaluation aims to evaluate the quality of overall explanation.

\begin{figure}[h]
\scriptsize
\centering
\begin{tcolorbox}
\textbf{\# Task Description}\\
You are a powerful multi-modal evaluation assistant tasked with evaluating explanation texts for questions related to generated images.\\
\\
\textbf{\# Input Data}\\
1. A question about a generated image. The explanation text should clarify the answer to this question.\\
2. An explanation text to be evaluated against the factual content of the image.\\
3. A reference explanation text, which correctly represents the image content and serves as the gold standard for evaluation.\\
\\
\textbf{\# Evaluation Guidelines}\\
Assign a score from 0 to 5, where a higher score indicates better alignment with the reference explanation:\\
- 0: The evaluated explanation contradicts the reference, is empty, or lacks relevant information.\\
- 1-2: The evaluated explanation shows poor relevance to the reference, contains insufficient information, or has many errors.\\
- 3-4: The evaluated explanation generally aligns with the reference but may miss some details or contain minor errors.\\
- 5: The evaluated explanation fully aligns with the reference, potentially providing richer information with minimal or no errors.\\
\\
\textbf{\# Precautions}\\
Focus on the factual content conveyed by the reference explanation. Ignore any statements such as 'the answer' or 'ground truth' if they appear.\\
\\
\textbf{\# Question}\\
\textcolor{red}{\{question\}}\\
\\
\textbf{\# Explanation to be Evaluated}\\
\textcolor{red}{\{gt\_exp\}}\\
\\
\textbf{\# Reference Explanation}\\
\textcolor{red}{\{ref\_exp\}}\\
\\
\textbf{\# Output Instructions}\\
Provide only one line as the output: the score as an integer value.\\
\\
Do not include any additional information beyond the score.
\end{tcolorbox}
\caption{Prompt template for subjective evaluation of fine-grained explanations.}
\label{fig:sub_eval_fine}
\end{figure}

\begin{figure}[h]
\scriptsize
\centering
\begin{tcolorbox}
\textbf{\# Task Description}\\
You are a powerful multi-modal evaluation assistant tasked with evaluating explanation texts for the quality of generated images.\\
\\
\textbf{\# Input Data}\\
1. A list of questions about a generated image, reflecting multiple aspects of the image.\\
2. Ground truth answers and explanations for each question, strictly based on the image content, serving as reference for your evaluation.\\
3. Explanation to be evaluated, where you assess consistency with the reference and whether it fully covers the provided information.\\
\\
\textbf{\# Evaluation Guidelines}\\
Assign a score from 0 to 5, where a higher score indicates better alignment with the reference explanation:\\
- 0: The evaluated explanation contradicts the reference, is empty, or lacks relevant information.\\
- 1-2: The evaluated explanation shows poor relevance to the reference, contains insufficient information, or has many errors.\\
- 3-4: The evaluated explanation generally aligns with the reference but may miss some details or contain minor errors.\\
- 5: The evaluated explanation fully aligns with the reference, potentially providing richer information with minimal or no errors.\\
\\
\textbf{\# Precautions}\\
Focus on the factual content conveyed by the reference explanation. Ignore any statements such as 'the answer' or 'ground truth' if they appear.\\
\\
\textbf{\# Questions and Reference Explanation}\\
\textcolor{red}{\{ref\_exp\}}\\
\\
\textbf{\# Explanation to be Evaluated}\\
\textcolor{red}{\{gt\_exp\}}\\
\\
\textbf{\# Output Instructions}\\
Provide only one line as the output: the score as an integer value.\\
\\
Do not include any additional information beyond the score.
\end{tcolorbox}
\caption{Prompt template for subjective evaluation of coarse-grained explanations.}
\label{fig:sub_eval_coarse}
\end{figure}

\section{Case Study\label{app:case_study}}
\begin{figure*}
    \centering
    \includegraphics[width=1.0\linewidth]{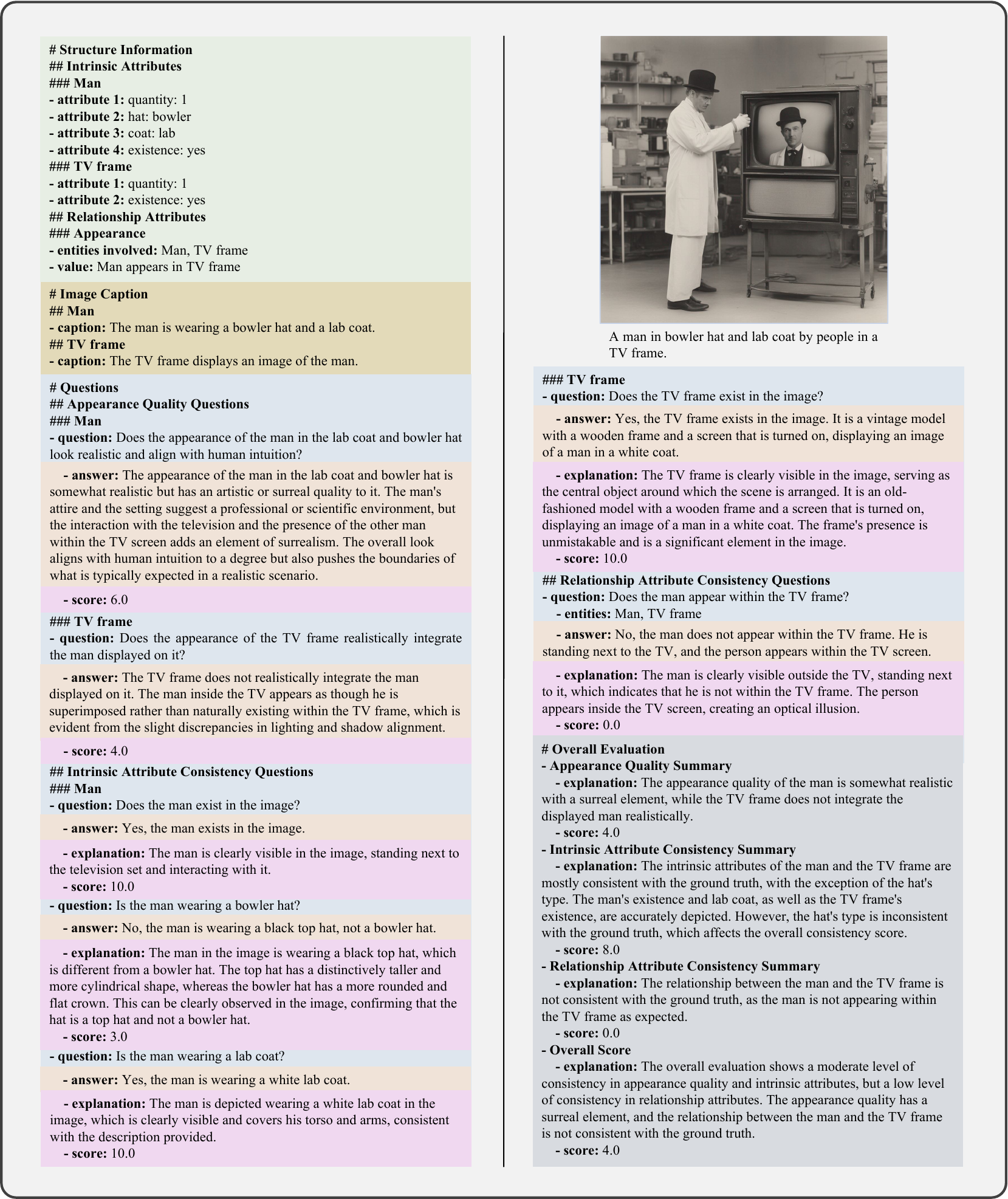}
    \caption{A case of the evaluation framework for open-source MLLMs.}
    \label{fig:case_study_coarse}
\end{figure*}

\begin{figure*}[h]
    \centering
    \includegraphics[width=1.0\linewidth]{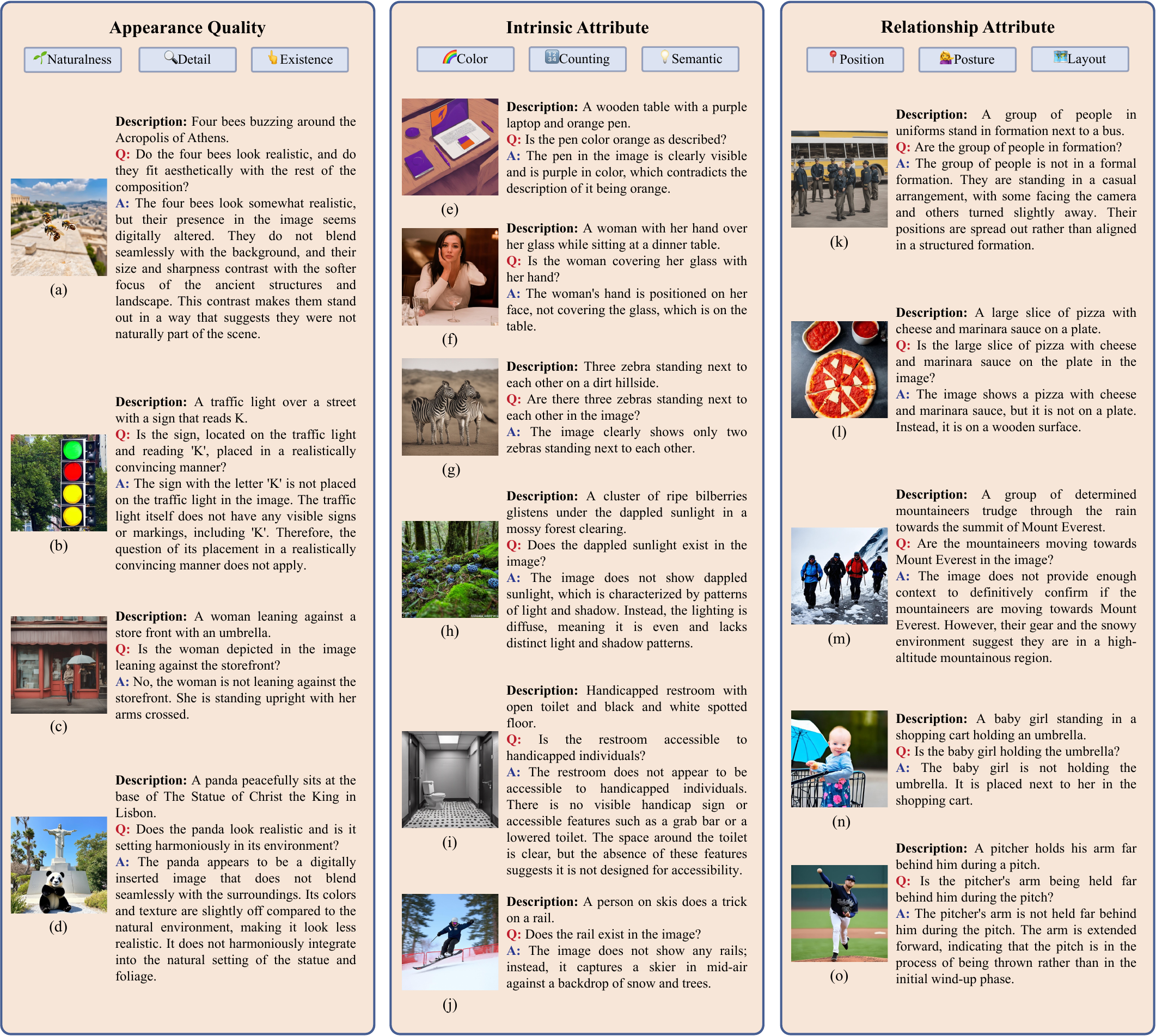}
    \caption{Cases for fine-grained evaluations in three categories.}
    \label{fig:case_study}
\end{figure*}

We provide an undivided case of evaluation with our proposed framework for open-source MLLMs in Figure~\ref{fig:case_study_coarse} and several individual questions in three categories (Appearance Quality, Intrinsic Attribute and Relationship Attribute) in Figure~\ref{fig:case_study}.

\input{sec/eval_prompt}

%% file: sec/eval_prompt.tex
\section{Evaluation Prompt Templates\label{app:framework}}
All prompt templates used in our proposed Task Decomposition Evaluation Framework are illustrated in Figure~\ref{fig:prompt_extract}, ~\ref{fig:prompt_answer} and \ref{fig:prompt_exp_score}.

\begin{figure*}[h]
\scriptsize
\centering
\begin{tcolorbox}[sidebyside]
\textbf{\# Your Task}\\
You are an expert in information extraction. Your task is to extract attributes of entities and relationships between entities from the text, and to pose a question about each entity's attributes and relationships.\\
\\
\textbf{\# Input Data}\\
The text is: \textcolor{red}{\{text\_prompt\}}\\
\\
\textbf{\# Extraction Pipeline}\\
\textbf{\#\# Step 1: Identify Entities}\\
\text{\quad} Step 1.1: Extract All Names\\
\text{\quad\quad} Extract all potential names from the input text.\\
\text{\quad} Step 1.2: Evaluate Each Name\\
\text{\quad\quad} - Determine Entity Status: For each extracted name, assess whether it qualifies as an entity based on context and predefined criteria.\\
\text{\quad\quad} - Include or Exclude: If a name is deemed an entity, include it in the output; otherwise, exclude it.\\
\\
\textbf{\#\# Step 2: Formulate a Question for Each Entity}\\
\text{\quad} For each entity, create a critical question regarding the realism, aesthetic appeal, and alignment with human intuition of the entity's appearance in the generated image. Focus questions primarily on overall authenticity rather than getting into detailed specifics.\\
\\
\textbf{\#\# Step 3: Identify All Attributes for Each Entity}\\
\text{\quad} Step 3.1: Identify Intrinsic Attributes\\
\text{\quad\quad} Intrinsic attributes are properties explicitly mentioned in the input text, such as color, size, shape, material, and quantity.\\
\text{\quad\quad} Step 3.1.1: Extract Quantity Attributes\\
\text{\quad\quad\quad} Identify words indicating quantity, including articles like ``a'' and ``an'', which suggest a quantity of one. For example, in ``a cat'', ``a'' indicates one cat. Attribute this quantity to the relevant entity.\\
\text{\quad\quad} Step 3.1.2: Extract Other Intrinsic Attributes\\
\text{\quad\quad\quad} Analyze words in the input text related to the entity, excluding the entity's name itself. Determine if these words denote intrinsic attributes and identify their types (e.g., color, size, material) and values.\\
\text{\quad\quad} Step 3.1.3: Verify Attribute Type and Value Pair\\
\text{\quad\quad\quad} Ignore attribute pairs if the value doesn't appear in the text, is part of the entity's name, or is ``unspecified''.\\
\text{\quad\quad} Step 3.1.4: Exclude Positional Attributes\\
\text{\quad\quad\quad} Disregard attributes related to position, orientation, distance, or location.\\
\text{\quad\quad} Step 3.1.5: Add Existence Attribute\\
\text{\quad\quad\quad} For each entity, add an attribute ``existence'' with a value of ``yes'' to indicate it should exist in the image.\\
\text{\quad\quad} Step 3.1.6: Default Unspecified Quantities\\
\text{\quad\quad\quad} If the text doesn't specify a quantity, set it to ``unspecified''.\\
\text{\quad\quad} Step 3.1.7: Consolidate and Output Attributes\\
\text{\quad\quad\quad} Add verified attribute type-value pairs to the output. Ensure all entities are included.\\
\text{\quad} Step 3.2: Identify Relationship Attributes\\
\text{\quad\quad} Relationship attributes describe an entity's relationship with other entities.\\
\text{\quad\quad} Step 3.2.1: Analyze Relation Words\\
\text{\quad\quad\quad} Identify words in the input text that describe relationships between entities, specifying the relationship type and related entities.\\
\text{\quad\quad} Step 3.2.2: Output Relationship Types\\
\text{\quad\quad\quad} Add identified relationships and related entities to the output.\\
\\
\textbf{\#\# Step 4: Construct Questions Based on Extracted Attributes}\\
\text{\quad} Step 4.1: Construct Intrinsic Attribute Consistency Questions\\
\text{\quad\quad} Step 4.1.1: Existence Questions\\
\text{\quad\quad\quad} Generate questions such as, ``Does the [entity] exist in the image?'' where [entity] is the entity's name.\\

\tcblower

\text{\quad\quad} Step 4.1.2: Attribute Value Questions\\
\text{\quad\quad\quad} Create a question for each intrinsic attribute pair about the attribute value of the entity.\\
\text{\quad\quad} Step 4.1.3: Verify the Number of Questions\\
\text{\quad\quad\quad} Ensure the number of questions equals the total number of intrinsic attribute-value pairs, including one existence and one quantity question for each entity.\\
\text{\quad} Step 4.2: Construct Relationship Attribute Consistency Questions\\
\text{\quad\quad} Step 4.2.1: Relationship Questions\\
\text{\quad\quad\quad} For each relational attribute of each entity, formulate a question about its value in relation to other entities.\\
\text{\quad\quad} Step 4.2.2: Ensure Coverage\\
\text{\quad\quad\quad} Ensure the number of questions matches the number of relationship attribute pairs, with each pair corresponding to one question.\\
\\
\textbf{\# Output Template}\\
Replace variables in `\{\{\}\}'\\
And if the text is like ``Three apples'', the entity should be ``apple'', and the attribute should be ``three''. Instead of ``apple 1, apple 2, apple 3'' as the entities.\\
Please generate your extracted structured information based on the following markdown template (Do NOT generate // comment in the template):\\
\\
\textbf{\# Structure Information}\\
\textbf{\#\# Intrinsic Attributes}\\
\#\#\# \{\{entity\}\}\\
- attribute 1: \{\{attribute 1 type\}\}: \{\{attribute 1 value\}\}\\
- attribute 2: \{\{attribute 2 type\}\}: \{\{attribute 2 value\}\}\\
- attribute 3: {{attribute 3 type}}: {{attribute 3 value}}\\
...\\
\#\#\# \{\{next entity or group\}\}\\
...\\
\\
\textbf{\#\# Relationship Attributes}\\
\#\#\# \{\{relationship attribute 1\}\}\\
- entities involved: \{\{entity 1, entity 2, ...\}\}\\
- value: \{\{relationship attribute value\}\}\\
\#\#\# \{\{next relationship attribute\}\}\\
...\\
\\
\textbf{\# Questions}\\
\textbf{\#\# Appearance Quality Questions}\\
\#\#\# \{\{entity 1 name\}\}\\
- question: \{\{entity 1 appearance quality question \}\}\\
\#\#\# \{\{next entity\}\}\\
...\\
\\
\textbf{\#\# Intrinsic Attribute Consistency Questions}\\
\#\#\# \{\{entity 1 name\}\}\\
- question 1: \{\{entity 1 intrinsic attribute consistency question 1\}\}\\
- question 2: \{\{entity 1 intrinsic attribute consistency question 2\}\}\\
- question 3: \{\{entity 1 intrinsic attribute consistency question 3\}\}\\
- question 4: \{\{entity 1 intrinsic attribute consistency question 4\}\}\\
- next question\\
...\\
\#\#\# \{\{next entity\}\}\\
...\\
\\
\textbf{\#\# Relationship Attribute Consistency Questions}\\
- question 1: \{\{relationship attribute consistency question 1\}\}\\
\text{\quad} - entities: \{\{entity 1\}\} \{\{entity 2\}\}\\
- question 2: \{\{relationship attribute consistency question 2\}\}\\
...\\
\end{tcolorbox}
\caption{Prompt template for evaluation content extraction.}
\label{fig:prompt_extract}
\end{figure*}

\begin{figure*}[h]
\scriptsize
\centering
\begin{tcolorbox}[sidebyside]
\textbf{\# Your Task}\\
You are an assistant specialized in answering questions based on the content of images.\\
\\
\textbf{\# Input Data}\\
\textbf{1. Question Input:} These are the questions you are to answer. They consist of three parts: appearance quality questions, intrinsic attribute consistency questions, and relationship attribute consistency questions. The questions are: \textcolor{red}{\{questions\}}\\
\textbf{2. Target Image:} Use this image to answer the questions.\\
\textbf{3. Reference Image:} Use this as a reference for authenticity when answering questions about appearance quality based on the target image.\\
\\
\textbf{\# Answer Pipeline}\\
\textbf{\#\# Step 1: Generate the Target Image Caption}\\
- Identify all entities in the target image.\\
- For each entity, generate a caption that includes the entity’s name and all attributes.\\
- Generate a caption for each entity that includes its name and all relationships.\\
These captions are solely for answering the intrinsic attribute consistency questions. If an entity in the image caption does not have those questions, ignore it.\\
\\
\textbf{\#\# Step 2: Answer the Appearance Quality Questions}\\
- For each question, identify if the entity is present in the target image. If present, proceed to the next step; if absent, assign a score of 0.\\
- For each appearance quality question, determine if the entity's appearance in the target image is realistic, aesthetically pleasing, and aligns with human intuition.\\
- Use the reference image for authenticity when needed.\\
- Assign a score from 0 to 10 for each question, and provide a brief explanation for the score awarded.\\
- \textbf{Scoring Strategy:}\\
\text{\quad} - 0-3: The appearance lacks realism, is not aesthetically pleasing, or does not align with human intuition.\\
\text{\quad} - 4-7: The appearance is somewhat realistic, aesthetically pleasing, or aligns with human intuition.\\
\text{\quad} - 8-10: The appearance is very realistic, aesthetically pleasing, and aligns well with human intuition.\\
\\
\textbf{\#\# Step 3: Answer the Intrinsic Attribute Consistency Questions}\\
- For each question, check if the entity exists in the target image. If it does, proceed; if not, state that the entity doesn't exist in the image.\\
- Answer each intrinsic attribute consistency question by detailing the corresponding attribute value from both the target image and its caption. Be thorough in your explanations; avoid simple yes or no answers.\\
\\
\textbf{Note:} You must address all questions in the question input.\\

\tcblower

\textbf{\#\# Step 4: Answer the Relationship Attribute Consistency Questions}\\
- For each question, verify the entity's presence in the target image. If present, continue; otherwise, indicate that the entity does not exist in the image.\\
- Determine the relationships of each entity in the target image and its caption. Provide a detailed answer, avoiding yes or no responses, and explain your reasoning.\\
\\
\textbf{\# Output Template}\\
Replace variables in `\{\{\}\}'\\
Please generate your result based on following markdown template (Do NOT generate // comment in the template).\\
\\
\textbf{\# Image Caption}\\
\#\# \{\{entity 1 name\}\}\\
- caption: \{\{entity 1 caption\}\}\\
\#\# \{\{next entity\}\}\\
...\\
\\
\textbf{\# Answers}\\
\textbf{\#\# Appearance Quality Questions}\\
\#\#\# \{\{entity 1 name\}\}\\
- question: \{\{entity 1 appearance quality question\}\}\\
\text{\quad} - explanation: \{\{explanation\}\}\\
\text{\quad} - score: \{\{score\}\}\\
\#\#\# \{\{next entity\}\}\\
...\\
\\
\textbf{\#\# Intrinsic Attribute Consistency Questions}\\
\#\#\# \{\{entity 1 name\}\}\\
- question 1: \{\{entity 1 intrinsic attribute consistency question 1\}\}\\
\text{\quad} - answer: \{\{answer\}\}\\
- question 2: \{\{entity 1 intrinsic attribute consistency question 2\}\}\\
\text{\quad} - answer: \{\{answer\}\}\\
- next question\\
...\\
\#\#\# \{\{next entity\}\}\\
...\\
\\
\textbf{\#\# Relationship Attribute Consistency Questions}\\
- question 1: \{\{relationship attribute consistency question 1\}\}\\
\text{\quad} - entities: \{\{entity 1\}\}, \{\{entity 2\}\}\\
\text{\quad} - answer: \{\{answer\}\}\\
- question 2: \{\{relationship attribute consistency question 2\}\}\\
\text{\quad} - entities: \{\{entity 1\}\}, \{\{entity 2\}\}\\
\text{\quad} - answer: \{\{answer\}\}\\
...\\
\end{tcolorbox}
\caption{Prompt template for caption and answer generation.}
\label{fig:prompt_answer}
\end{figure*}

\begin{figure*}[h]
\scriptsize
\centering
\begin{tcolorbox}[sidebyside]
\textbf{\# Your Task}\\
You are an expert in assessing the similarity between answers obtained from images and ground truth obtained from text.\\
\\
\textbf{\# Input Data}\\
\textbf{1. Answers from the Image:} These are the answers you need to evaluate including three components:\\
\text{\quad} - Appearance Quality Answers\\
\text{\quad} - Intrinsic Attribute Consistency Answers\\
\text{\quad} - Relationship Attribute Consistency Answers\\
\text{\quad} The provided answer is: \textcolor{red}{\{answer\}}\\
\\
\textbf{2. Ground Truth:} This is the standard to which you will compare the image answers. It consists of two components:\\
\text{\quad} - Entities' Attributes\\
\text{\quad} - Relationships\\
\text{\quad} The structured information is the sole ground truth: \textcolor{red}{\{{structure\_info\}}}\\
\\
\textbf{\# Scoring Strategy}\\
- 0-3: The answer is not consistent with the ground truth at all.\\
- 4-7: The answer is somewhat consistent with the ground truth; semantics are similar but not entirely aligned.\\
- 8-10: The answer is very consistent with the ground truth.\\
\\
\textbf{\# Evaluation Pipeline}\\
\textbf{\#\# Step 1: Evaluate Appearance Quality Answers}\\
- Focus solely on the appearance quality of the answers.\\
\\
\textbf{\#\# Step 2: Evaluate Intrinsic Attribute Consistency Answers}\\
- For each intrinsic attribute consistency answer of every entity, compare it with the corresponding ground truth.\\
- If the entity does not appear in the image, assign a score of 0. Otherwise, proceed to the next step.\\
- Offer a short explanation of how well the answer matches the ground truth.\\
- Provide a score reflecting the extent of the match; if there is no match, score it as zero. In cases of mismatch, assign the lowest possible score.\\
\\
\textbf{\#\# Step 3: Evaluate Relationship Attribute Consistency Answers}\\
- For each relationship's attribute consistency answer, compare it with the ground truth.\\
- If the entity does not exist in the image, assign a score of 0. Otherwise, proceed to the next step.\\
- Offer a short explanation of how well the answer matches the ground truth.\\
- Provide a score reflecting the extent of the match; if there is no match, score it as zero. In cases of mismatch, assign the lowest possible score.\\
\\
\textbf{\#\# Step 4: Overall Evaluation}\\
- Combine your findings on appearance quality, summarize your observations, and assign a score based on this summary.\\
- Summarize the degree of match between the image answers and the intrinsic attribute consistency of the ground truth, and assign a score based on this evaluation.\\
- Summarize the degree of match for relationship attribute consistency between the image answers and the ground truth, and assign a score based on this summary.\\

\tcblower

- Integrate all summaries regarding appearance quality, intrinsic attribute consistency, and relationship attribute consistency. Offer a comprehensive evaluation description and assign a final score based on this description.\\
\\
\textbf{\# Output Template}\\
Replace Variable in `\{\{\}\}'\\
Please generate your output based on following markdown template (Do NOT generate // comment in the template).\\

\textbf{\# Evaluation}\\
\textbf{\#\# Appearance Quality Answers}\\
\#\#\# \{\{entity 1 name\}\}\\
- question: \{\{entity 1 appearance quality question\}\}\\
\text{\quad} - explanation: \{\{explanation\}\}\\
\text{\quad} - score: \{\{score\}\}\\
\#\#\# \{\{next entity\}\}\\
...\\
\\
\textbf{\#\# Intrinsic Attribute Consistency Answers}\\
\#\#\# \{\{entity 1 name\}\}\\
- question 1: \{\{entity 1 intrinsic attribute consistency question 1\}\}\\
\text{\quad} - answer: \{\{answer from the image\}\}\\
\text{\quad} - explanation: \{\{explanation\}\}\\
\text{\quad} - score: \{\{score\}\}\\
- question 2: \{\{entity 1 intrinsic attribute consistency question 2\}\}\\
\text{\quad} - answer: \{\{answer from the image\}\}\\
\text{\quad} - explanation: \{\{explanation\}\}\\
\text{\quad} - score: \{\{score\}\}\\
- next question\\
...\\
\#\#\# \{\{next entity\}\}\\
...\\
\\
\textbf{\#\# Relationship Attribute Consistency Answers}\\
- question 1: \{\{relationship attribute consistency question 1\}\}\\
\text{\quad} - entities: \{\{entity 1\}\} \{\{entity 2\}\}\\
\text{\quad} - answer: \{\{answer from the image\}\}\\
\text{\quad} - explanation: \{\{explanation\}\}\\
\text{\quad} - score: \{\{score\}\}\\
- question 2: \{\{relationship attribute consistency question 2\}\}\\
...\\
\\
\textbf{\#\# Overall Evaluation}\\
- Appearance Quality Summary:\\
\text{\quad} - explanation: \{\{explanation\}\}\\
\text{\quad} - score: \{\{score\}\}\\
- Intrinsic Attribute Consistency Summary:\\
\text{\quad} - explanation: \{\{explanation\}\}\\
\text{\quad} - score: \{\{score\}\}\\
- Relationship Attribute Consistency Summary:\\
\text{\quad} - explanation: \{\{explanation\}\}\\
\text{\quad} - score: \{\{score\}\}\\
- Overall Score:\\
\text{\quad} - explanation: \{\{explanation\}\}\\
\text{\quad} - score: \{\{score\}\}\\
\end{tcolorbox}
\caption{Prompt template for explanation and scoring.}
\label{fig:prompt_exp_score}
\end{figure*}